\pdfoutput=1

\documentclass[11pt]{article}

\usepackage[final]{EMNLP2023}

\usepackage{times}
\usepackage{latexsym}

\usepackage[T1]{fontenc}

\usepackage[utf8]{inputenc}

\usepackage{microtype}

\usepackage{inconsolata}

\usepackage{graphicx}
\usepackage{subfig}
\usepackage{float} 
\usepackage{overpic}
\usepackage{amsmath}

\usepackage{xcolor}
\usepackage{color}
\usepackage{colortbl}

\usepackage{booktabs}
\usepackage{multirow}
\usepackage[normalem]{ulem}
\useunder{\uline}{\ul}{}
\usepackage{booktabs}

\usepackage{hyperref}
\makeatletter
\def\UrlAlphabet{%
      \do\a\do\b\do\c\do\d\do\e\do\f\do\g\do\h\do\i\do\j%
      \do\k\do\l\do\m\do\n\do\o\do\p\do\q\do\r\do\s\do\t%
      \do\u\do\v\do\w\do\x\do\y\do\z\do\A\do\B\do\C\do\D%
      \do\E\do\F\do\G\do\H\do\I\do\J\do\K\do\L\do\M\do\N%
      \do\O\do\P\do\Q\do\R\do\S\do\T\do\U\do\V\do\W\do\X%
      \do\Y\do\Z}
\def\UrlDigits{\do\1\do\2\do\3\do\4\do\5\do\6\do\7\do\8\do\9\do\0}
\g@addto@macro{\UrlBreaks}{\UrlOrds}
\g@addto@macro{\UrlBreaks}{\UrlAlphabet}
\g@addto@macro{\UrlBreaks}{\UrlDigits}
\makeatother

\title{
Skill-Based Few-Shot Selection for In-Context Learning
}

\author{
  Shengnan An\thanks{\quad Work done during the internship at Microsoft.}\hspace{0.4mm} $^{\diamondsuit,\clubsuit}$,\, Bo Zhou$^{*\heartsuit,\clubsuit}$,\, Zeqi Lin\thanks{\quad Corresponding authors.}\hspace{0.4mm} $^{\clubsuit}$,\, Qiang Fu$^{\clubsuit}$,\, Bei Chen$^{\clubsuit}$, \\
  \textbf{Nanning Zheng$^{\dagger\diamondsuit}$,\, Weizhu Chen$^{\clubsuit}$,\, Jian-Guang LOU$^{\clubsuit}$}\\
  $^{\diamondsuit}$National Key Laboratory of Human-Machine Hybrid Augmented Intelligence, \\National Engineering Research Center of Visual Information and Applications, \\Institute of Artificial Intelligence and Robotics, Xi'an Jiaotong University\\
  $^{\clubsuit}$Microsoft Corporation,\, 
  $^{\heartsuit}$Northeastern University\\
  $^{\diamondsuit}$\texttt{\{an1006634493@stu, nnzheng@mail\}.xjtu.edu.cn, $^{\heartsuit}$2171386@stu.neu.edu.cn} \\
  $^{\clubsuit}$\texttt{\{Zeqi.Lin, qifu, beichen, wzchen, jlou\}@microsoft.com}
}

\begin{document}
\maketitle
\begin{abstract}
\textit{In-context learning} is the paradigm that adapts large language models to downstream tasks by providing a few examples.
\textit{Few-shot selection}---selecting appropriate examples for each test instance separately---is important for in-context learning.
In this paper, we propose \textsc{Skill-KNN}, a skill-based few-shot selection method for in-context learning.
The key advantages of \textsc{Skill-KNN} include: (1) it addresses the problem that existing methods based on pre-trained embeddings can be easily biased by surface natural language features that are not important for the target task; (2) it does not require training or fine-tuning of any models, making it suitable for frequently expanding or changing example banks.
The key insight is to optimize the inputs fed into the embedding model, rather than tuning the model itself.
Technically, \textsc{Skill-KNN} generates the skill-based descriptions for each test case and candidate example by utilizing a pre-processing few-shot prompting, thus eliminating unimportant surface features.
Experimental results across five cross-domain semantic parsing datasets and six backbone models show that \textsc{Skill-KNN} significantly outperforms existing methods.
\end{abstract}

\section{Introduction}\label{sec:intro}

\begin{figure}[t]
	\centering
	\subfloat[Raw-Input-Based Selection\label{fig:surface_form_retrieving}
]{\includegraphics[width=.9\linewidth]{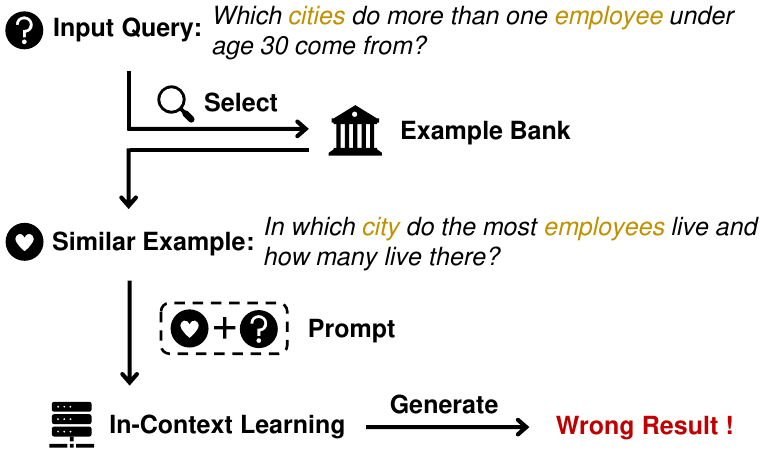}}\\
	\subfloat[Skill-Based Selection\label{fig:selection_with_skill}]{\includegraphics[width=.9\linewidth]{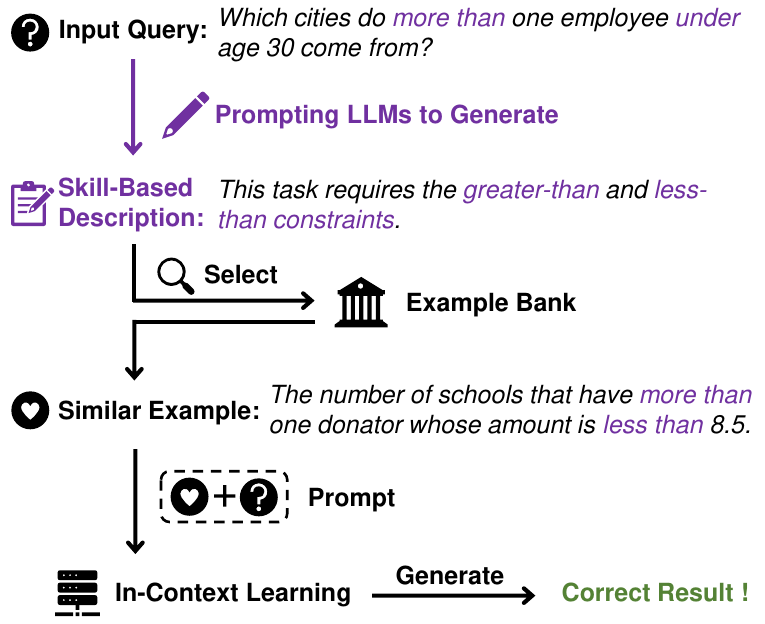}}
	\caption{In-context learning with different selection methods.
    (a) Examples from raw-input-based selection just share similar entities with the input query.
    (b) With the skill-based description, the selected examples contain the desired task-specific skills.
    }
\end{figure}

\begin{figure*}[t]
    \centering
    \includegraphics[width=.95\textwidth]{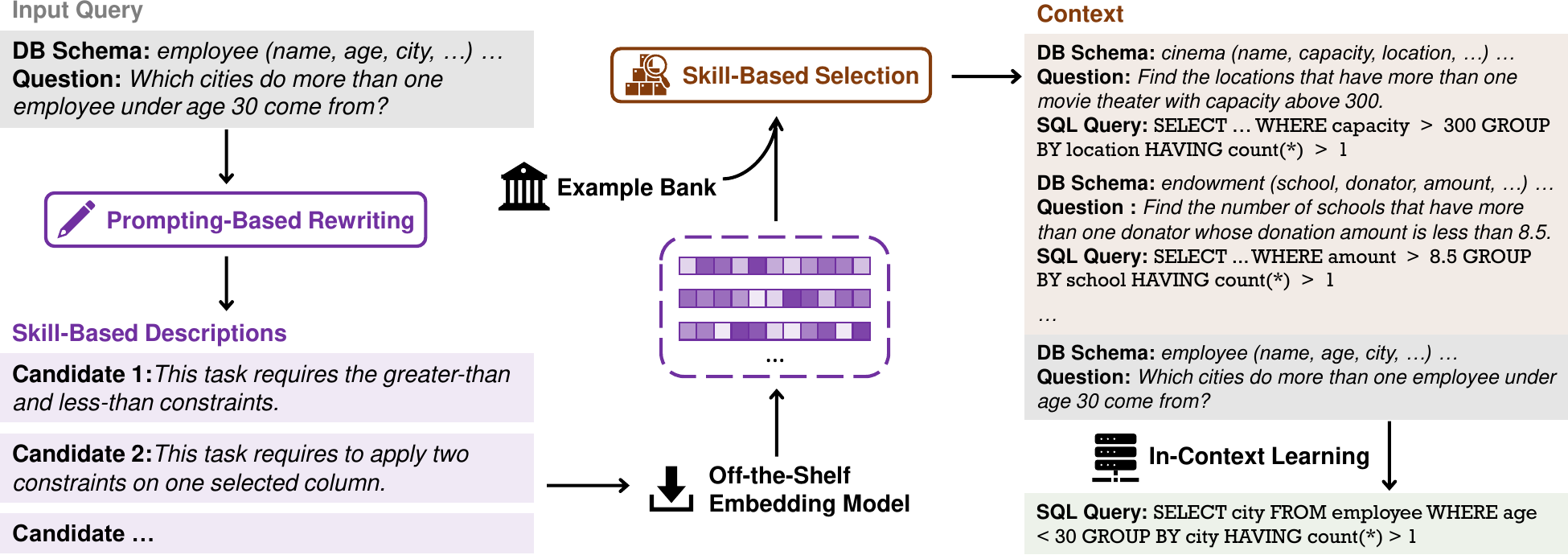}
    \caption{
    The bird's-eye view of \textsc{Skill-KNN}, a rewrite-then-retrieve selection method to facilitate in-context learning with skill-based descriptions.
    }
    \label{fig:overall}
\end{figure*}

In-context learning~\citep{brown2020language} has become a prevailing paradigm for utilizing large language models (LLMs)~\citep{hendrycks2020measuring, patel2021mapping, rae2021scaling, zhang2022opt, hoffmann2022training, srivastava2022beyond, chowdhery2022palm, smith2022using, wei2022emergent}.
It employs a frozen task-agnostic backbone model to serve various downstream tasks without requiring parameter updates for each task.
Under in-context learning, the LLMs generate output for an input query by conditioning on the prompt that contains input-output examples.
Due to limited context length of the language model, only a few examples can be presented in the prompt.
Prior studies have found that the performance of in-context learning is sensitive to the selected in-context examples~\citep{liu2022makes, zhang2022active, chen2023relation}.
Therefore, one essential research question is: \textit{how to select proper examples from a large example bank?}

\textit{Raw-input-based selection} is one widely applied solution~\citep{gao2021making, liu2022makes, hu2022context}.
It involves embedding raw inputs of examples using an off-the-shelf embedding model and then selecting the most similar examples.
It can be conveniently applied in various downstream tasks.
However, this method can be easily biased by surface natural language features that are not important for the target task.
For instance, in semantic parsing tasks\footnote{Semantic parsing means to parse an NL utterance into a machine-understandable logical form (e.g., a SQL query).}, the raw-input-based selection just finds out examples with similar entities (as illustrated in Figure~\ref{fig:surface_form_retrieving}), while the better in-context examples should contain the required executable operations in logical forms, which can be regarded as the \textit{task-specific skills}.

\begin{table*}[t]
\caption{Part of our annotated skill-based descriptions for text-to-SQL tasks.
}
\label{tab:annotated_skills}
\centering
\resizebox{.99\linewidth}{!}{
\begin{tabular}{@{}lll@{}}
\toprule
Input Query & Database Schema & Skill-Based Description \\ \midrule
Show all majors. & \begin{tabular}[c]{@{}l@{}}allergy type [allergy, allergytype]\\ has allergy [stuid, allergy]\\ student [stuid, lname, fname, age, sex, major, ...]\end{tabular} & \begin{tabular}[c]{@{}l@{}}To solve this task in the database, we need to select distinct \\ values in the column.\end{tabular} \\ \midrule
\begin{tabular}[c]{@{}l@{}}Count the number of different colleges that players \\ who play for Columbus Crew are from.\end{tabular} & \begin{tabular}[c]{@{}l@{}}team [team id, name]\\ country [country id, country name, capital, ...]\\ match season [season, player, position, country, team, ...] ...\end{tabular} & \begin{tabular}[c]{@{}l@{}}To solve this task in the database, we need to join two tables \\ and count the number of distinct values in the column.\end{tabular} \\ \midrule
\begin{tabular}[c]{@{}l@{}}Which catalog contents have a product stock number \\ that starts from "2"? Show the catalog entry names.\end{tabular} & \begin{tabular}[c]{@{}l@{}}catalogs [catalog id, catalog name, catalog publisher, ...]\\ catalog structure [catalog level number, catalog id, ...]\\ catalog contents [catalog entry id, catalog level number, ...] ...\end{tabular} & \begin{tabular}[c]{@{}l@{}}To solve this task in the database, we need to select one column \\ and apply a constraint on the format of values in this column.\end{tabular} \\ \bottomrule
\end{tabular}
}
\end{table*}

To overcome this limitation,
we aim to make this embedding-based selection better aware of the intrinsic skills behind the raw inputs.
We consider to harness the power of \textbf{prompting LLMs to convert the desired skills from raw inputs}, which maintains the training-free advantage during selection.
There has been much work trying to fine-tune the embedding model for each task based on the example bank~\citep{rubin2022learning, poesia2022synchromesh, hu2022context, ye2023compositional}.
However, fine-tuning-based methods are difficult to apply in practical scenarios:
it is laborious to train and save the embedding model for each task, and it is also inconvenient to re-train the model on a dynamic example bank that can be updated frequently.

Specifically, we introduce \textbf{\textsc{Skill-KNN}}, a training-free, skill-based selection method (briefly illustrated in Figure~\ref{fig:selection_with_skill}).
Overall, 
\textsc{Skill-KNN} will first \textbf{generate skill-based descriptions from raw input queries},
then feed these descriptions into an off-the-shelf embedding model to select most similar examples.
To generate skill-based descriptions, we prompt a frozen LLM with just a few human-annotated demonstrations, 
which does not require any fine-tuning process and has no rule-based constraint.
Additionally, to alleviate the sensitivity to the order of annotated demonstrations during generation, we design two variants of \textsc{Skill-KNN}:
we sample a set of candidate descriptions by shuffling annotated demonstrations, then select candidate based on \textbf{consistency} and \textbf{distinctiveness}, respectively.

The experimental results show that \textsc{Skill-KNN} brings a considerable boost for in-context learning compared to the raw-input-based selection.
We evaluate \textsc{Skill-KNN} on five challenging semantic parsing datasets: Spider~\citep{yu2018spider}, Dr. Spider~\citep{chang2023drspider}, KaggleDBQA~\citep{lee2021kaggledbqa}, BIRD~\citep{li2023llm}, and COGS~\citep{kim2020cogs}.
We take six models for in-context learning: text-chat-davinci-002, code-davinci-002, text-davinci-003, code-cushman-002, gpt-35-turbo, and gpt-4.
Across these tasks and models, \textsc{Skill-KNN} consistently performs best among non-oracle selection methods and, at times, is even comparable to oracle methods.
For instance, with text-chat-davinci-002, \textsc{Skill-KNN} achieves 78.3\% execution accuracy on Spider, while the best raw-input-based selection method reaches 74.6\% and Target-KNN\footnote{One of the oracle methods, detailed in Section~\ref{sec:selection_methods}.} attains 78.6\%.
Furthermore, our ablation study indicates that \textsc{Skill-KNN} retains its superiority when constraints are imposed on the annotated demonstrations,  including reducing the number of demonstrations, restricting the database diversity, and decreasing the operation coverage.

Our contributions are three-fold:
1) we propose a skill-based few-shot selection method \textsc{Skill-KNN}, which leverages the power of prompting LLMs to generate skill-based descriptions;
2) we design two variants of \textsc{Skill-KNN} based on consistency and distinctiveness, respectively;
3) our comprehensive experiments across various semantic parsing tasks and backbone models demonstrate the effectiveness of \textsc{Skill-KNN}, and our analysis of annotated demonstrations provides further insights for better utilization of \textsc{Skill-KNN}.

\section{Preliminaries}\label{sec:background}

In this section, we introduce embedding-based few-shot selection as the preliminary of our method.

\subsection{In-Context Learning with Few-Shot Selection}

Consider a downstream task $\mathcal{T}$ that contains a set of input-output examples $\{(x_{i}\rightarrow{y_{i}})\}^{n}$ (termed \textit{example bank} $\mathcal{B}$)\footnote{In semantic parsing tasks, each input query contains a natural language question along with the database schema.} , and a pre-trained large language model with frozen parameters $\theta$.
Given a test input query $x_t$, the large language model with in-context learning will generate an output $y_t$ by sampling from the following distribution,
\begin{equation}\label{sec:icl}
    y_t \sim \mathrm{LLM}_{\theta,\tau}[R(x_t,\mathcal{B})\oplus x_t],
\end{equation}
in which $\tau$ is the sampling temperature, $R(x_t,\mathcal{B})$ returns a sequence of examples selected from $\mathcal{B}$ according to $x_t$, and $\oplus$ means to sequentially concatenate two sequences.
In the later part of the paper, we omit the frozen $\theta$ and set $\tau=0$ by default, which means to perform greedy decoding.

\textit{Few-shot selection} aims to design the algorithm $R(x_t,\mathcal{B})$ that can work well for task $\mathcal{T}$.

\subsection{Embedding-Based Few-Shot Selection}

A standard implementation of $R(x_t,\mathcal{B})$ is to leverage an off-the-shelf embedding model $\mathrm{Emb}(\cdot)$ and calculate the embedding similarity of raw inputs~\citep{liu2022makes},
\begin{equation}\label{equ:cosine_input}
    \mathrm{sim}(x_t, x_i) = \frac{\mathrm{Emb}(x_t)\mathrm{Emb}(x_i)^{T}}{|\mathrm{Emb}(x_t)||\mathrm{Emb}(x_i)|},
\end{equation}
in which $x_i$ is the input\footnote{For raw-input-based selection, we only use the question in the input query for embedding and omit the database schema, as it contains much trivial and redundant information for the question and could confuse the embedding model.} of one example $(x_i, y_i)\in\mathcal{B}$.
Based on Equation~\ref{equ:cosine_input}, we can select $k$ most similar examples from $\mathcal{B}$.
In addition, these examples will be sorted in the prompt according to their similarities to the test input query: examples with higher similarity scores will be placed closer to the test input query.

This standard implementation of $R(x_t,\mathcal{B})$ is a raw-input-based selection.
It just searches for examples with similar inputs (i.e., the $x_t$ and $x_i$ in Equation~\ref{equ:cosine_input}).
Some recent researches propose to fine-tune the embedding model (from $\mathrm{Emb}(\cdot)$ to $\mathrm{Emb}'(\cdot)$) ~\citep{rubin2022learning, poesia2022synchromesh, hu2022context, ye2023compositional}.
In this paper, we want to explore how to improve the effectiveness of few-shot selection without training or fine-tuning of any models.

\section{\textsc{Skill-KNN}}\label{sec:rewrite_then_retrieve}

\textsc{Skill-KNN} involves a rewrite-then-retrieve process to better exploit the potential of in-context learning.
Figure~\ref{fig:overall} gives a bird's-eye view of our method.
To mine and utilize task-specific skills, \textsc{Skill-KNN} contains a \textbf{prompting-based rewriting} stage and a \textbf{skill-based selection} stage.
Prompting-based rewriting will prompt LLMs to generate skill-based descriptions from the given input query.
Skill-based selection will return few-shot examples based on these generated descriptions.
In the following, we elaborate the design of \textsc{Skill-KNN}.

\subsection{Generating Skill-Based Descriptions}\label{sec:input_to_skill}

We prompt a frozen large language model to rewrite each input query as a skill-based description, which does not require any fine-tuning process.
Specifically, we first annotate the skill-based descriptions for 16 examples in $\mathcal{B}$, then prompt the large language model with these annotated demonstrations and generate for other examples in $\mathcal{B}$ and for each test input query.

Note that we annotated skills with natural language descriptions rather than rule-based constraints.
It is important to note that off-the-shelf embedding models are primarily pre-trained on natural language (NL) data and may not be well-suited for handling specifically designed structural constraints.
By annotating skills with NL descriptions, we can better align with the off-the-shelf embedding models, which in turn allows us to leverage their generalizability when encoding unannotated NL descriptions more effectively.
Thus, these natural language skills can better suit the off-the-shelf embedding model, and our annotated demonstrations can better generalize to unseen data.

Formally, with a set of annotated demonstrations $\{x_a\rightarrow s_a\}$ in which $s_a$ is the annotated skill-based description for the raw input $x_a$, we generate the $s_i$ for each unannotated input $x_i$ by prompting the large language model,
\begin{equation}\label{equ:generate_skill}
    s_i = \mathrm{LLM}[\{x_a\rightarrow s_a\}\oplus x_i],
\end{equation}
then, these descriptions are fed into the off-the-shelf embedding model to select similar examples,
\begin{equation}\label{equ:cosine_skill}
    \mathrm{sim}(s_t, s_i) = \frac{\mathrm{Emb}(s_t)\mathrm{Emb}(s_i)^{T}}{|\mathrm{Emb}(s_t)||\mathrm{Emb}(s_i)|}.
\end{equation}
Table~\ref{tab:annotated_skills} shows part of our annotated demonstrations for text-to-SQL tasks, and all our annotations are contained in Appendix~\ref{sec:ap_annotated_examples}.
Note that different text-to-SQL tasks can share the same set of annotated demonstrations in our experiments.

Equation~\ref{equ:cosine_skill} defines the basic version of \textsc{Skill-KNN}.
Moreover, we notice that the generated skill-based descriptions sometimes could be sensitive to the order of annotated demonstrations.
Such a sensitivity is also observed in some previous work~\citep{zhao2021calibrate, lu2022fantastically}.
Therefore, we design two variants of \textsc{Skill-KNN} to further address this sensitivity issue.

\subsection{Variants}\label{sec:skill_enhanced_selection}

\begin{figure}[t]
    \centering
    \includegraphics[width=.48\textwidth]{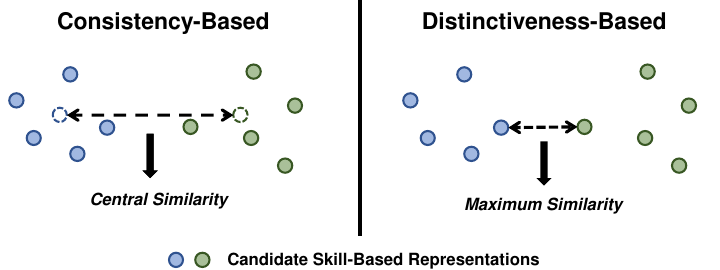}
    \caption{
    Two variants of \textsc{Skill-KNN}. The blue and green points represent two candidate sets of skill-based representations.
    }
    \label{fig:enhancing_methods}
\end{figure}

To alleviate the influence from the sensitivity to prompt order, we design two variants of \textsc{Skill-KNN} that change the order of annotated demonstrations and perform rewriting multiple times.
Specifically, for each input $x_i$, both two variants generate a set of candidate descriptions $S_i=\{s_i^1, s_i^2, ..., s_i^m\}$ according to Equation~\ref{equ:generate_skill} by changing the order in $\{x_a\rightarrow s_a\}$.
Then, two variants use these candidate descriptions from the view of consistency and distinctiveness, respectively.
Figure~\ref{fig:enhancing_methods} briefly illustrates the basic ideas behind these two variants.
Appendix~\ref{sec:ap_motivation} provides further analysis for the motivation behind the design of two variants.

\paragraph{Consistency-Based Variant.}
From the view of consistency, we take the central embedding of all candidate descriptions during selecting examples,
\begin{equation}\label{equ:enhance_consistency}
    \mathrm{sim}_c (S_t, S_i) = \frac{\overline{e_t}\,\overline{e_i}^{T}}{|\overline{e_t}|\,|\overline{e_i}|},\,
    \overline{e} = \frac{1}{m} \sum_{s^j\in S} \mathrm{Emb}(s^j),
\end{equation}
in which $S_t$ and $S_i$ represent two sets of candidate descriptions for the test input $x_t$ and one example $(x_i, y_i)\in\mathcal{B}$, respectively.
This variant is inspired by prior work on improving the consistency of chain-of-thought reasoning~\citep{wang2022self, li2022advance}.
As illustrated in the left in Figure~\ref{fig:enhancing_methods}, Equation~\ref{equ:enhance_consistency} can be regarded as an embedding-level majority voting among all candidate descriptions during selection.

\paragraph{Distinctiveness-Based Variant.}
Considering that the central embedding can sometimes be overwhelmed by trivial candidates,
we want to highlight the most distinctive and informative description among all candidates.
Formally, we consider the maximum similarity score between two sets for selection,
\begin{equation}\label{equ:enhance_distinctiveness}
    \mathrm{sim}_d (S_t, S_i) = \max_{j,k}\frac{\mathrm{Emb}(s_t^j)\mathrm{Emb}(s_i^k)^{T}}{|\mathrm{Emb}(s_t^j)||\mathrm{Emb}(s_i^k)|},
\end{equation}
in which $s_t^j\in S_t$ and $s_i^k\in S_i$.
As illustrated in the right in Figure~\ref{fig:enhancing_methods}, Equation~\ref{equ:enhance_distinctiveness} means that we take the minimum distance among two set of candidates for selecting similar examples.

\section{Experimental Setup}

\begin{table*}[t]
\caption{Our main experimental results (\%) across various LLMs and tasks.
Numbers in \textbf{bold} are the best results across non-oracle methods, and results with {\ul underlines} can outperform at least one oracle method.
}
\label{tab:main_results}
\centering
\resizebox{.92\linewidth}{!}{
\begin{tabular}{@{}p{4cm}<{\centering}p{7cm}<{}p{1.2cm}<{\centering}p{1.2cm}<{\centering}p{1.2cm}<{\centering}p{1.2cm}<{\centering}p{1.2cm}<{\centering}p{1.2cm}<{\centering}p{1.2cm}<{\centering}p{1.2cm}<{\centering}p{1.2cm}<{\centering}@{}}
\toprule
\multirow{2}{*}{Backbone} & \multicolumn{1}{c}{\multirow{2}{*}{Method}} & \multirow{2}{*}{Spider} & \multicolumn{3}{c}{Dr. Spider} & \multirow{2}{*}{KDBQA} & \multirow{2}{*}{BIRD} & \multicolumn{2}{c}{COGS} & \multirow{2}{*}{\textit{\#Wins}} \\ \cmidrule(lr){4-6} \cmidrule(lr){9-10}
 & \multicolumn{1}{c}{} &  & DB & NLQ & SQL &  &  & P.S. & P.A. &  \\ \midrule
\multirow{13}{*}{text-chat-davinci-002} & Random & 72.9 & 54.1 & 58.1 & 68.2 & 24.1 & 35.7 & 59.1 & 61.5 & 0 \\
 & KNN w/ SBERT~\citep{liu2022makes} & 73.0 & 51.6 & 58.2 & 67.3 & 24.3 & 38.3 & 78.5 & 71.1 & 0 \\
 & KNN w/ OpenAI   Babbage~\citep{liu2022makes} & 74.0 & 53.2 & 61.0 & 69.2 & 35.3 & 38.1 & 88.3 & 74.1 & 0 \\
 & KNN w/ OpenAI Ada~\citep{liu2022makes} & 72.8 & 51.8 & 59.2 & 69.6 & 31.9 & 37.1 & 81.3 & 64.2 & 0 \\
 & MMR w/ OpenAI   Ada~\citep{ye2022complementary} & 74.6 & 54.1 & 60.7 & 71.3 & 37.3 & 37.2 & 86.8 & 78.9 & 0 \\ \cmidrule(l){2-11} 
 & \textsc{Skill-KNN} w/ SBERT (base) & 76.8 & 55.3 & 60.3 & 72.1 & 34.7 & 38.9 & 93.8 & 88.3 & 0 \\
 & \quad + consistency & 76.0 & 54.8 & 60.3 & 71.9 & 33.8 & 38.2 & 94.3 & 87.9 & 0 \\
 & \quad + distinctiveness & \textbf{78.3} & {\ul \textbf{57.0}} & \textbf{61.4} & \textbf{72.2} & {\ul \textbf{40.0}} & 38.0 & 92.6 & 88.8 & 5 \\
 & \textsc{Skill-KNN} w/ OpenAI Ada (base) & 77.1 & 55.4 & 60.5 & 70.9 & 38.7 & \textbf{38.9} & 94.9 & 95.6 & 1 \\
 & \quad + consistency & 77.2 & 54.8 & 59.6 & 71.1 & {\ul \textbf{40.0}} & 38.3 & {\ul \textbf{95.2}} & 93.5 & 2 \\
 & \quad + distinctiveness & 77.2 & 56.5 & 59.8 & 70.1 & 39.5 & 38.3 & 93.3 & {\ul \textbf{97.0}} & 1 \\ \cmidrule(l){2-11} 
 & Target-KNN (oracle) & 78.6 & 56.6 & 65.5 & 75.5 & 37.8 & 41.7 & 86.8 & 83.2 & - \\
 & Target Sketch Matching   (oracle) & 79.5 & 54.2 & 65.1 & 76.2 & 40.4 & 40.8 & 96.9 & 95.7 & - \\ \midrule
\multirow{13}{*}{code-davinci-002} & Random & 74.2 & 53.9 & 59.3 & 70.2 & 27.9 & 38.4 & 56.1 & 63.1 & 0 \\
 & KNN w/ SBERT~\citep{liu2022makes} & 73.1 & 51.2 & 58.6 & 69.1 & 31.2 & 39.9 & 75.4 & 64.7 & 0 \\
 & KNN w/ OpenAI   Babbage~\citep{liu2022makes} & 73.6 & 50.7 & 59.6 & 69.1 & 37.5 & 38.2 & 85.4 & 72.4 & 0 \\
 & KNN w/ OpenAI Ada~\citep{liu2022makes} & 72.7 & 50.4 & 60.2 & 69.5 & 35.7 & 38.2 & 77.3 & 60.3 & 0 \\
 & MMR w/ OpenAI   Ada~\citep{ye2022complementary} & 74.8 & 53.7 & 60.7 & 69.4 & 37.8 & 37.9 & 84.9 & 75.4 & 0 \\ \cmidrule(l){2-11} 
 & \textsc{Skill-KNN} w/ SBERT (base) & 76.4 & 53.0 & 60.2 & \textbf{72.4} & 38.6 & 40.2 & 93.1 & 86.8 & 1 \\
 & \quad + consistency & 77.1 & 51.1 & 60.7 & \textbf{72.4} & 36.8 & 40.1 & 93.5 & 87.1 & 1 \\
 & \quad + distinctiveness & \textbf{77.4} & {\ul \textbf{55.0}} & 60.9 & 71.0 & \textbf{43.0} & 38.9 & 91.9 & 88.4 & 3 \\
 & \textsc{Skill-KNN} w/ OpenAI Ada (base) & 76.2 & 54.7 & \textbf{61.4} & 70.2 & 39.8 & 40.3 & 94.0 & 92.9 & 1 \\
 & \quad + consistency & 76.4 & 54.9 & 60.2 & 71.8 & 39.5 & \textbf{40.8} & {\ul \textbf{96.2}} & 88.8 & 2 \\
 & \quad + distinctiveness & 76.6 & {\ul \textbf{55.0}} & 60.6 & 70.2 & 38.9 & \textbf{40.8} & 93.5 & {\ul \textbf{94.0}} & 3 \\ \cmidrule(l){2-11} 
 & Target-KNN (oracle) & 78.8 & 56.1 & 68.2 & 76.4 & 44.5 & 44.3 & 85.9 & 79.7 & - \\
 & Target Sketch Matching   (oracle) & 80.9 & 54.0 & 66.7 & 77.9 & 44.9 & 43.8 & 95.0 & 94.8 & - \\ \midrule
\multirow{13}{*}{text-davinci-003} & Random & 69.0 & 52.2 & 55.1 & 64.5 & 20.6 & 36.0 & 65.5 & 61.2 & 0 \\
 & KNN w/ SBERT~\citep{liu2022makes} & 69.9 & 50.2 & 56.1 & 67.2 & 21.0 & 37.7 & 82.1 & 71.1 & 0 \\
 & KNN w/ OpenAI   Babbage~\citep{liu2022makes} & 72.2 & 53.4 & 58.0 & 69.5 & 26.8 & 37.7 & 88.8 & 77.2 & 0 \\
 & KNN w/ OpenAI Ada~\citep{liu2022makes} & 70.8 & 50.3 & 57.2 & 66.3 & 30.3 & 36.5 & 82.8 & 64.2 & 0 \\
 & MMR w/ OpenAI   Ada~\citep{ye2022complementary} & 72.4 & 53.3 & 58.6 & 69.9 & 31.4 & 37.7 & 87.1 & 81.0 & 0 \\ \cmidrule(l){2-11} 
 & \textsc{Skill-KNN} w/ SBERT (base) & 74.9 & 54.4 & 59.0 & 70.2 & 32.9 & 38.0 & 94.8 & 88.4 & 0 \\
 & \quad + consistency & 75.3 & 54.8 & 59.0 & 70.6 & 32.0 & 38.1 & 94.3 & 89.2 & 0 \\
 & \quad + distinctiveness & {\ul \textbf{76.6}} & 54.1 & 58.9 & 70.6 & {\ul \textbf{36.8}} & 37.4 & 93.3 & 87.5 & 2 \\
 & \textsc{Skill-KNN} w/ OpenAI Ada (base) & 74.2 & 55.2 & \textbf{59.5} & 68.5 & 34.0 & 38.4 & 95.7 & 94.3 & 1 \\
 & \quad + consistency & 74.3 & 55.0 & \textbf{59.5} & \textbf{71.0} & {\ul \textbf{36.8}} & {\ul \textbf{40.2}} & {\ul \textbf{96.7}} & 92.7 & 5 \\
 & \quad + distinctiveness & 73.7 & {\ul \textbf{56.2}} & 59.4 & 70.6 & 32.4 & 37.9 & 93.8 & {\ul \textbf{97.4}} & 2 \\ \cmidrule(l){2-11} 
 & Target-KNN (oracle) & 75.6 & 54.2 & 63.9 & 73.3 & 35.1 & 40.0 & 87.8 & 84.5 & - \\
 & Target Sketch Matching   (oracle) & 76.4 & 52.5 & 61.4 & 72.0 & 31.6 & 39.8 & 97.6 & 95.3 & - \\ \midrule
\multirow{13}{*}{code-cushman-002} & Random & 72.2 & 51.5 & 56.6 & 66.8 & 26.1 & 35.3 & 56.7 & 55.6 & 0 \\
 & KNN w/ SBERT~\citep{liu2022makes} & 67.6 & 47.8 & 54.4 & 64.6 & 29.4 & 37.0 & 70.3 & 62.5 & 0 \\
 & KNN w/ OpenAI   Babbage~\citep{liu2022makes} & 71.6 & 48.6 & 56.4 & 66.8 & 36.4 & 36.4 & 77.5 & 71.6 & 0 \\
 & KNN w/ OpenAI Ada~\citep{liu2022makes} & 68.9 & 48.1 & 57.3 & 67.4 & 31.9 & 34.8 & 71.3 & 54.7 & 0 \\
 & MMR w/ OpenAI   Ada~\citep{ye2022complementary} & 69.3 & 49.8 & 57.1 & 68.6 & 36.2 & 36.8 & 75.6 & 72.8 & 0 \\ \cmidrule(l){2-11} 
 & \textsc{Skill-KNN} w/ SBERT (base) & 74.5 & 52.1 & 58.0 & 69.3 & 35.1 & 37.0 & 90.6 & 84.1 & 0 \\
 & \quad + consistency & \textbf{74.7} & 52.3 & 58.0 & 69.6 & 35.3 & 38.3 & 91.1 & 86.6 & 1 \\
 & \quad + distinctiveness & 72.8 & 52.0 & \textbf{58.8} & 67.2 & {\ul \textbf{40.8}} & 35.9 & 91.8 & 83.2 & 2 \\
 & \textsc{Skill-KNN} w/ OpenAI Ada (base) & 73.6 & 52.4 & 58.4 & 69.0 & 38.4 & 37.9 & 93.2 & 86.6 & 0 \\
 & \quad + consistency & 73.5 & {\ul \textbf{53.0}} & 57.9 & \textbf{69.8} & 40.0 & \textbf{38.6} & {\ul \textbf{95.0}} & 82.8 & 4 \\
 & \quad + distinctiveness & 73.7 & 51.2 & 57.6 & 67.9 & 38.9 & 37.3 & 91.9 & {\ul \textbf{90.9}} & 1 \\ \cmidrule(l){2-11} 
 & Target-KNN (oracle) & 77.6 & 52.5 & 65.6 & 73.8 & 40.8 & 41.7 & 77.0 & 72.8 & - \\
 & Target Sketch Matching   (oracle) & 77.2 & 51.7 & 62.2 & 73.2 & 39.7 & 39.8 & 91.9 & 94.0 & - \\ \bottomrule
\end{tabular}
}
\end{table*}

\begin{table}[t]
\caption{Performance of gpt-35-turbo and gpt-4 on Spider dev set.
}
\label{tab:35_4}
\centering
\resizebox{.99\linewidth}{!}{
\begin{tabular}{@{}clc@{}}
\toprule
Backbone & Method & Exec. Acc. \\ \midrule
\multirow{6}{*}{gpt-4} & Random & 76.1 \\ 
 & KNN w/ SBERT~\citep{liu2022makes} & 76.7 \\
 & Din-SQL~\citep{pourreza2023dinsql} & 74.2 \\ \cmidrule(l){2-3} 
 & \textsc{Skill-KNN} w/ SBERT (base) & 81.3 \\ 
 & \quad + consistency & \textbf{82.7} \\
 & \quad + distinctiveness & 82.3 \\ \midrule
\multirow{6}{*}{gpt-35-turbo} & Random & 74.3 \\ 
 & KNN w/ SBERT~\citep{liu2022makes} & 73.7 \\
 & KNN w/ OpenAI Babbage~\citep{liu2022makes} & 73.4 \\ \cmidrule(l){2-3} 
 & \textsc{Skill-KNN} w/ SBERT (base) & 76.2 \\ 
 & \quad + consistency & 76.3 \\
 & \quad + distinctiveness & \textbf{76.8} \\ \bottomrule
\end{tabular}
}
\end{table}

\begin{table}[t]
\caption{Comparison with the fine-tuning-based selection methods on Spider dev set.
}
\label{tab:fine_tune}
\centering
\resizebox{.99\linewidth}{!}{
\begin{tabular}{@{}clc@{}}
\toprule
Backbone & Method & Exec. Acc. \\ \midrule
\multirow{4}{*}{text-chat-davinci-002} & \textsc{EPR}~\citep{rubin2022learning} & 74.4 \\ 
 & \textsc{CEIL}~\citep{ye2023compositional} & 75.0 \\ 
 & \textsc{TST}~\citep{poesia2022synchromesh} & 76.3 \\ \cmidrule(l){2-3} 
 & \textsc{Skill-KNN} w/ SBERT (base) & 76.8 \\ \midrule
\multirow{4}{*}{code-davinci-002} & \textsc{EPR}~\citep{rubin2022learning} & 75.9 \\ \cmidrule(l){2-3} 
 & \textsc{Skill-KNN} w/ SBERT (base) & 76.4 \\
 & \quad + consistency & 77.1 \\
 & \quad + distinctiveness & \textbf{77.4} \\ \midrule
\multirow{4}{*}{text-davinci-003} & \textsc{EPR}~\citep{rubin2022learning} & 69.7 \\ \cmidrule(l){2-3} 
 & \textsc{Skill-KNN} w/ SBERT (base) & 74.9 \\
 & \quad + consistency & 75.3 \\
 & \quad + distinctiveness & \textbf{76.6} \\ \midrule
\multirow{4}{*}{code-cushman-002} & \textsc{EPR}~\citep{rubin2022learning} & 74.6 \\ \cmidrule(l){2-3} 
 & \textsc{Skill-KNN} w/ SBERT (base) & 74.5 \\
 & \quad + consistency & \textbf{74.7} \\
 & \quad + distinctiveness & 72.8 \\ \bottomrule
\end{tabular}
}
\end{table}

\begin{table}[t]
\caption{Performance of \textsc{Skill-KNN} and two baselines on GSM8K.
}
\label{tab:gsm8k}
\centering
\resizebox{.99\linewidth}{!}{
\begin{tabular}{@{}clc@{}}
\toprule
Backbone & Method & Accuracy \\ \midrule
\multirow{3}{*}{text-chat-davinci-002} & Random & 69.1 \\ 
 & KNN w/ SBERT~\citep{liu2022makes} & 69.9 \\ \cmidrule(l){2-3} 
 & \textsc{Skill-KNN} w/ SBERT (base) & 71.0 \\ \bottomrule
\end{tabular}
}
\end{table}

In this section, we will introduce the tasks, compared selection methods, backbone models, and hyper-parameters in our experiments.

\subsection{Tasks}

We evaluate on five challenging cross-domain semantic parsing datasets.
Due to the cross-domain property, the model can not easily solve these tasks by just copying some similar surface features from the provided in-context examples.

\vspace{2mm}
\noindent \textbf{Spider}~\citep{yu2018spider} is a large-scale text-to-SQL dataset.
It contains a train set with 7,000 examples and a dev set with 1,034 examples.
Moreover, the train set and dev set do not share any database.
We take the train set of Spider as the example bank, and evaluate on the dev set.

\vspace{2mm}
\noindent \textbf{Dr. Spider}~\citep{chang2023drspider} is a diagnostic evaluation benchmark constructed based on Spider.
It contains 15,269 examples which can be divided into 3 sub-tasks, according to the type of designed perturbations: database perturbations (DB), natural language question perturbations (NLQ), and SQL query perturbations (SQL).
We take the train set of Spider as the example bank, since Dr. Spider is purely an evaluation benchmark.

\vspace{2mm}
\noindent \textbf{KaggleDBQA}~\citep{lee2021kaggledbqa} (KDBQA) is a small while complex dataset towards realistic evaluation of text-to-SQL semantic parsers.
It contains 8 real Web databases with original formatting and 275 unrestricted questions.
Since ther is not much data in KDBQA, we take it as a pure test set and use the train set of Spider as the example bank.

\vspace{2mm}
\noindent \textbf{BIRD}~\citep{li2023llm} is a large scale text-to-SQL dataset with real-world database contents.
It has 9,428 training examples and 1,534 test cases in the dev set.
Each case in BIRD is equipped with a description of the required external knowledge, which is not contained in above three text-to-SQL tasks.
Since the database schema in BIRD is too large, we first take grounding to reduce the size of schema (detailed in Appendix~\ref{sec:ap_bird}).

\vspace{2mm}
\noindent \textbf{COGS}~\citep{kim2020cogs} is a synthetic benchmark for testing compositional generalization in semantic parsing.
It can also be regarded as a cross-domain setting, containing a significant distribution shift between train set and test set.
The logical form in COGS represents the thematic roles in the input query (detailed in Appendix~\ref{sec:ap_input_output_formats}).
We use the output format designed in \citet{an2023incontext} and evaluate on two sub-tasks, primitive substitution (P.S.) and primitive structural alternation (P.A.).

\begin{figure*}[t]
	\centering
    \includegraphics[width=.24\linewidth]{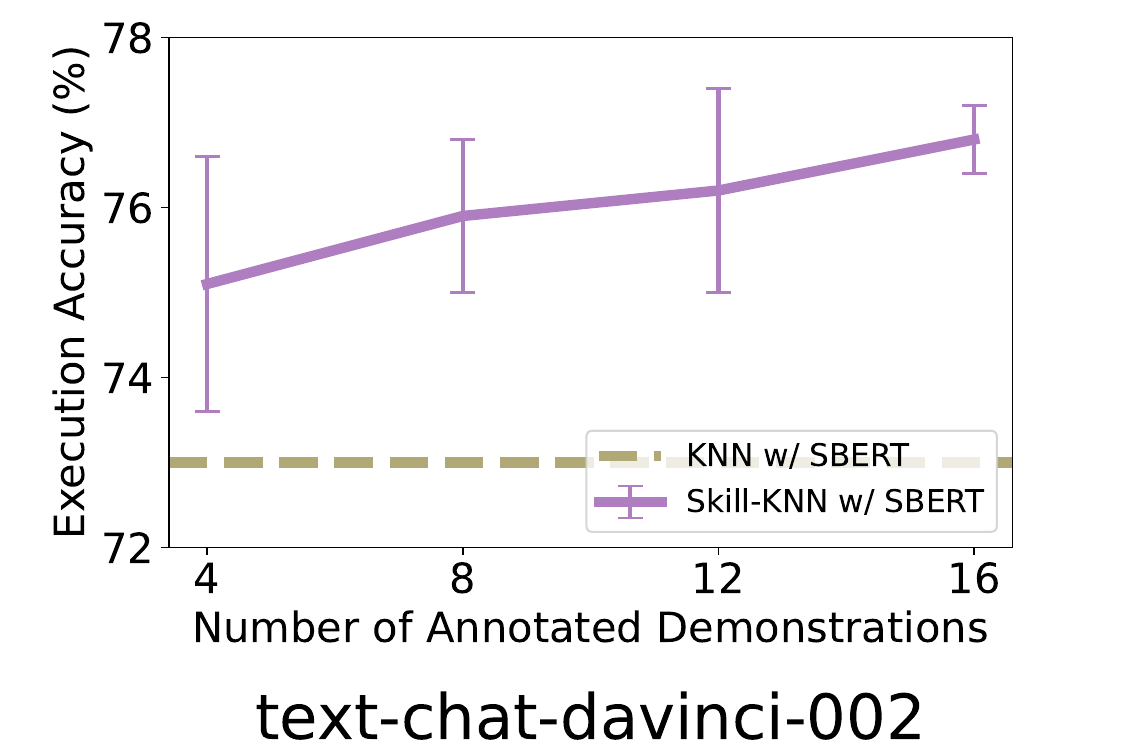}
    \includegraphics[width=.24\linewidth]{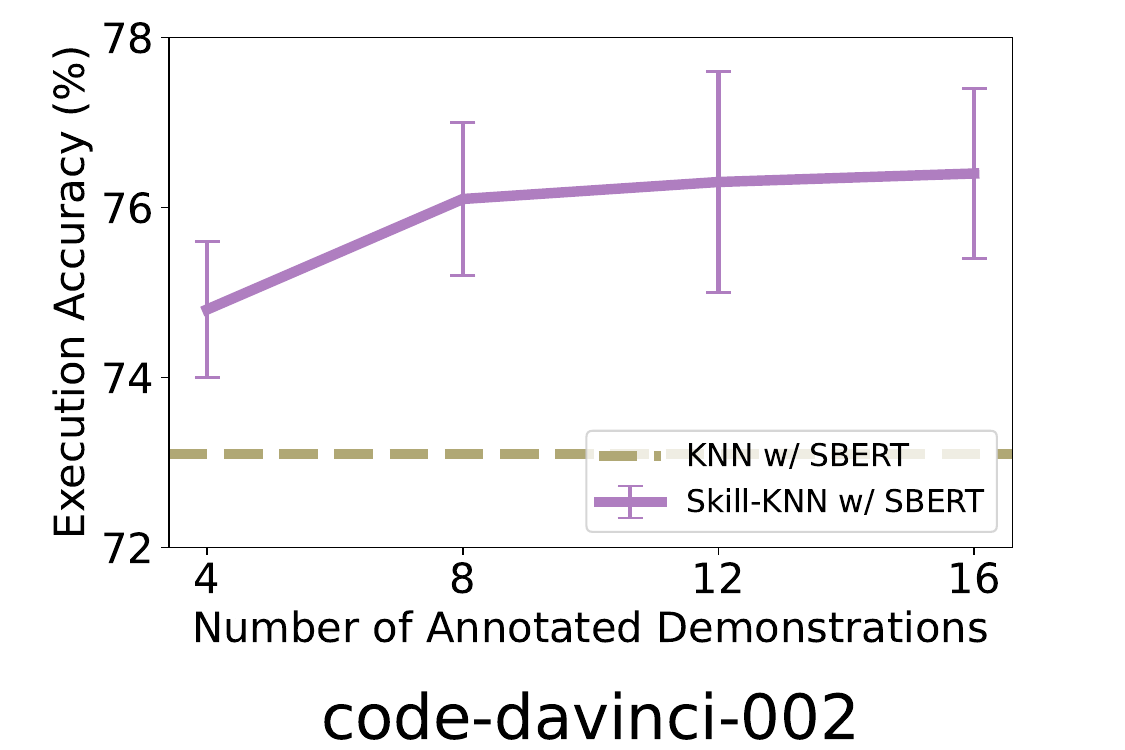}
	\includegraphics[width=.24\linewidth]{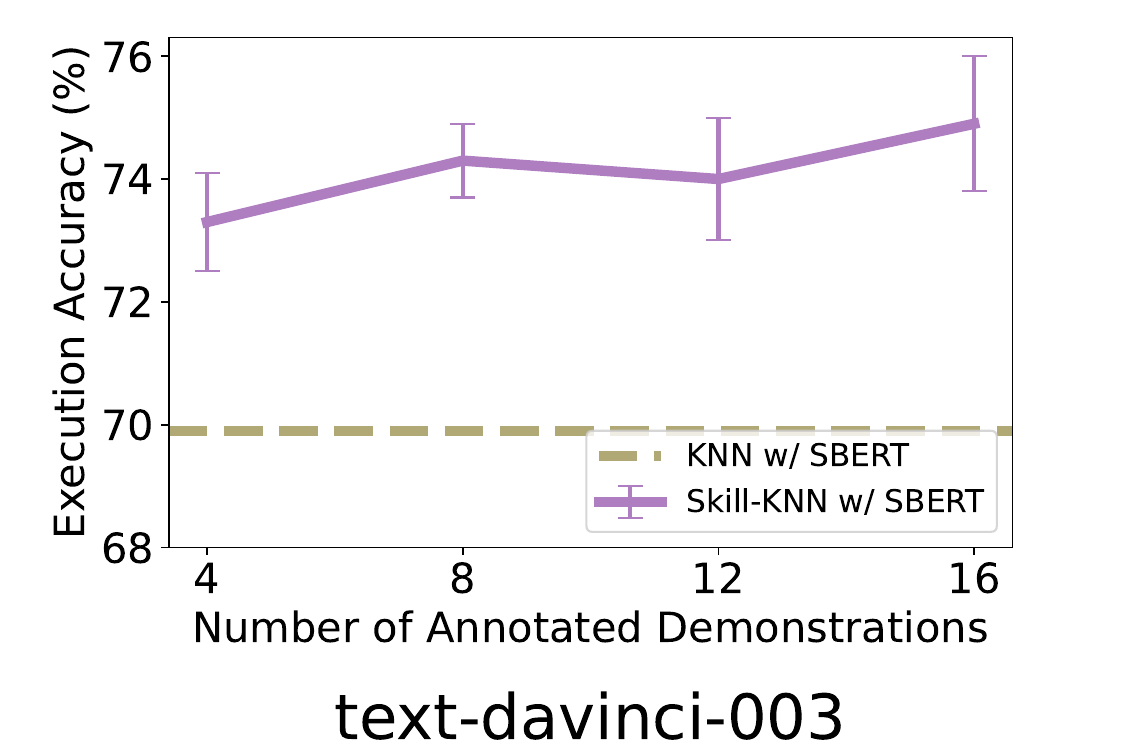}
    \includegraphics[width=.24\linewidth]{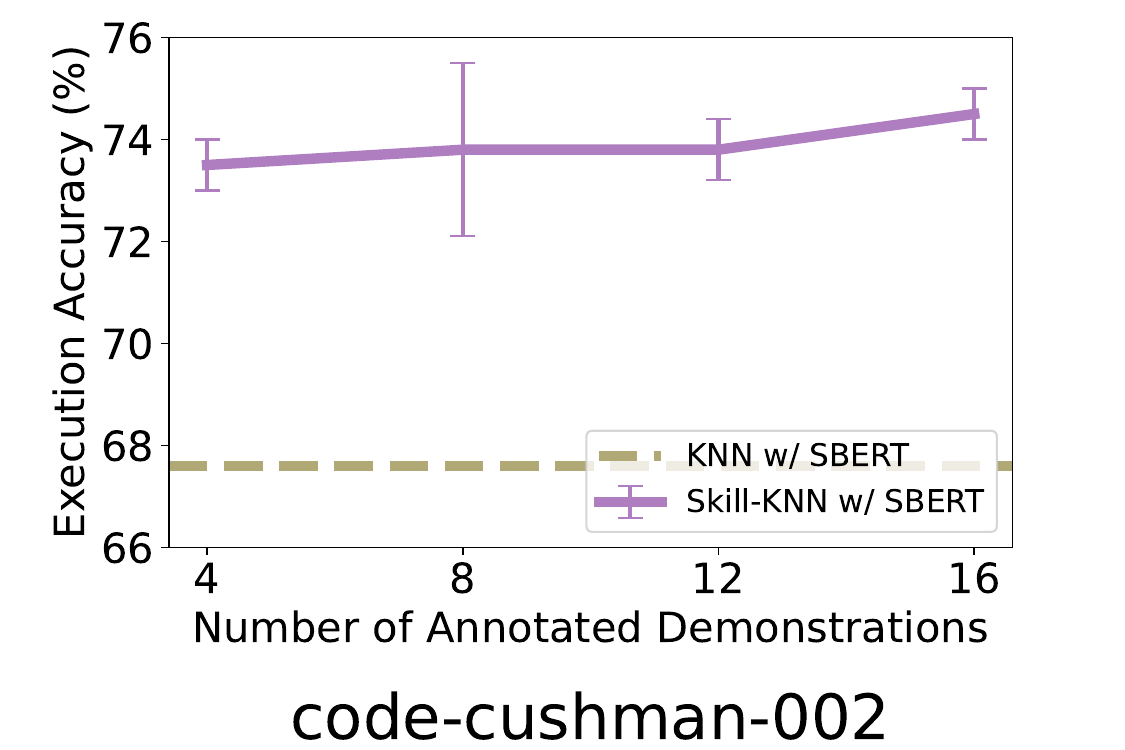}
	\caption{Performance of \textsc{Skill-KNN} (base version) with different number of annotated demonstrations.
    }\label{fig:analysis_length}
\end{figure*}

\subsection{Selection Methods}\label{sec:selection_methods}
We mainly compare \textsc{Skill-KNN} with training-free selection methods.

\vspace{2mm}
\noindent \textbf{Random.}
We randomly select examples from $\mathcal{B}$ as in-context examples.
For each test case, we take random selections 3 times and average the results.

\vspace{2mm}
\noindent \textbf{KNN}~\citep{liu2022makes}.
We test three off-the-shelf embedding models:
Sentence-BERT (SBERT) with all-mpnet-base-v2 checkpoint\footnote{\url{https://www.sbert.net/}.}~\cite{reimers2019sentence}, OpenAI embedding model\footnote{\url{https://platform.openai.com/docs/guides/embeddings/what-are-embeddings}.} with text-similarity-babbage-001 checkpoint (OpenAI Babbage), and OpenAI embedding model with text-embedding-ada-002 checkpoint (OpenAI Ada).
KNN with OpenAI embedding models can serve as strong baselines for training-free selection methods, as these large models have been well pre-trained for judging text similarity~\citep{neelakantan2022text}.

\vspace{2mm}
\noindent \textbf{MMR}~\citep{ye2022complementary} is a dynamic selection method to enhance the diversity of selected examples from KNN.
It adds a penalty term according to the similarity to the already selected examples.
We take OpenAI Ada for embedding and follow the implementation details in \citet{ye2022complementary}.

\vspace{2mm}
\noindent \textbf{\textsc{Skill-KNN} (ours).}
We test \textsc{Skill-KNN} with SBERT and OpenAI Ada.
For the base version of \textsc{Skill-KNN} (i.e., without consistency or distinctiveness), we shuffle the order of annotated demonstrations to generate $m=5$ skill-based descriptions for each input query and average the results.
There is a balance between achieving optimal performance and minimizing computational costs.
We provide more experimental analysis in Appendix~\ref{sec:ap_m}.
For two variants, we take all 5 generated descriptions as the candidate set.

\vspace{2mm}
Besides these training-free baselines, we also compare with three fine-tuning-based methods.
These methods leverage the example bank for fine-tuning the embedding model.

\vspace{2mm}
\noindent \textbf{\textsc{EPR}}~\citep{rubin2022learning} requires a scoring LM to produce positive/negative examples for fine-tuning the embedding model, and we use GPT-J~\citep{gpt-j}, a 6B-parameter LM as the scoring LM\footnote{OpenAI language models cannot be used as the scoring function since OpenAI API does not provide this functionality.}.

\vspace{2mm}
\noindent \textbf{\textsc{CEIL}}~\citep{ye2023compositional} proposes a compositional selection method for tn-context
learning.
It models the compositional interaction between the given input and in-context examples, and fine-tunes the selection model through a carefully-designed contrastive learning objective.

\vspace{2mm}
\noindent \textbf{\textsc{TST}}~\citep{poesia2022synchromesh} retrieves few-shot examples from a training bank using target similarity tuning.
It learns to recognize utterances that describe similar target programs despite differences in surface natural
language features.

\vspace{2mm}
In addition, we also compare with two oracle methods, in which ground truth output sequences are allowed to be leveraged for few-shot selection.

\vspace{2mm}
\noindent \textbf{Target-KNN (oracle).}
We select examples with similar output embeddings.
We use OpenAI Babbage and OpenAI Ada to encode the ground truth, and take the best result of two models for each task.

\vspace{2mm}
\noindent \textbf{Target Sketch Matching (oracle).}
We select in-context examples with similar sketches of ground truth.
For text-to-SQL tasks, we calculate the overlap of SQL key words (detailed in Appendix~\ref{sec:ap_target_sketch_match_sql}).
For COGS, we follow the target-side structural similarity setting in \citet{an2023incontext}.

\subsection{Backbones and Hyper-parameters}

We conduct experiments with six OpenAI language models as the backbones\footnote{In this work, the ``backbone'' refers to the frozen large language model for prompting.}:
text-chat-davinci-002, code-davinci-002, text-davinci-003, code-cushman-002, gpt-35-turbo, and gpt-4.
For generating skill-based descriptions, we always use gpt-3.5-turbo, as it is cheap and fast.
We select $k=4$ in-context examples in all experiments.
We use execution-with-values accuracy\footnote{We use the official evaluation scripts for Spider in \href{https://github.com/taoyds/test-suite-sql-eval}{https://github.com/taoyds/test-suite-sql-eval}.} as the evaluation metric for text-to-SQL tasks and exact-match accuracy for COGS.

\begin{figure*}[t]
	\centering
    \includegraphics[width=.99\linewidth]{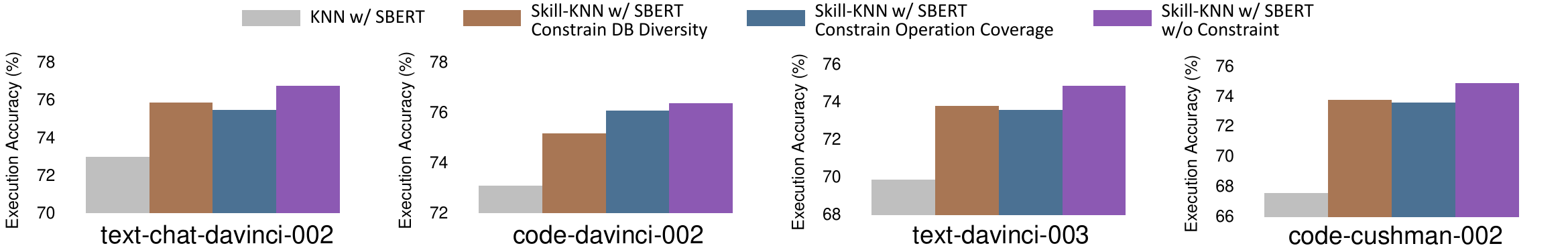}
	\caption{Performance of \textsc{Skill-KNN} (base version) with constraints on selecting examples for annotating.
    }\label{fig:analysis_selection}
\end{figure*}

\section{Main Results}\label{sec:main_results}

Table~\ref{tab:main_results}, Table~\ref{tab:35_4}, and Table~\ref{tab:fine_tune} report the main experimental results.
We also count the number of wins, i.e., how many tasks (and sub-tasks) the method performs best on.

\paragraph{\textsc{Skill-KNN} performs better than raw-input-based selection methods.}
Across all backbone models and tasks, our skill-based selections achieve the best performance among non-oracle methods.
Especially, \textsc{Skill-KNN} with SBERT can even outperform KNN with OpenAI embedding models.
These results clearly demonstrates the necessity and effectiveness of our prompting-based rewriting.
Appendix~\ref{sec:ap_more_compare} contains more experimental comparisons with existing selection methods.

\paragraph{\textsc{Skill-KNN} performs comparable/better than fine-tuning-based method.}
Results in Table~\ref{tab:fine_tune} show that \textsc{Skill-KNN} can perform comparable or even better than fine-tuning-based methods.
It demonstrates that optimizing the input to the embedding model can also effectively help downstream tasks without any fine-tuning.

\paragraph{Variants with consistency and distinctiveness can outperform the base version of \textsc{Skill-KNN}.}
As shown in the \textit{\#Wins} column in Table~\ref{tab:main_results}, two variants of \textsc{Skill-KNN} outperform the base version in most situations.
It demonstrates that injecting consistency and distinctiveness can effectively alleviate the order sensitivity.
Overall, we recommend choosing the distinctiveness variant as it wins more times than the consistency variant (19 vs 15).
When looking into detailed results,
different models prefer different variants of \textsc{Skill-KNN}:
text-chat-davinci-002 and code-davinci-002 prefer to the distinctiveness variant, while text-davinci-003 and code-cushman-002 prefer to the consistency variant.
We identify the potential factor that could lead to different preferences in Appendix~\ref{sec:analysis_two_enhance}.

\paragraph{\textsc{Skill-KNN} is more robust to perturbations than raw-input-based selections.}
Results on Dr. Spider reflect the robustness towards perturbations in data.
For instance, with text-chat-davinci-002, KNN with SBERT performs lower than the random baseline on two out of three types of perturbations, while all three versions of \textsc{Skill-KNN} outperform the random baseline on all three perturbations.
It indicates that \textsc{Skill-KNN} leads to more robust in-context learning than raw-input-based methods.

\paragraph{\textsc{Skill-KNN} can be effective in the math reasoning task.}
Our study primarily evaluated the effectiveness of \textsc{Skill-KNN} in the semantic parsing/code generation field.
To further examine the generalizability of \textsc{Skill-KNN} beyond semantic parsing, we have applied it to a challenging math reasoning task, GSM8K~\citep{Cobbe2021TrainingVT}.
Results in Table~\ref{tab:gsm8k} evidence that \textsc{Skill-KNN} can also be effective in tasks beyond semantic parsing.

\section{Analysis}
The most important mechanism in \textsc{Skill-KNN} is the prompting-based rewriting which requires a few manually annotated demonstrations.
Here we investigate how would these demonstrations affect the performance of \textsc{Skill-KNN}.
We experimentally analyze two factors:
the number of annotated demonstrations and the selection of examples for annotation.
The following analysis takes the base version of \textsc{Skill-KNN} with SBERT as the embedding model and the Spider dataset for evaluation.

\paragraph{More annotated demonstrations bring marginal improvements.}
We decrease the default number of annotated demonstrations from 16 to 4/8/12.
The results depicted in Figure~\ref{fig:analysis_length} illustrate that a gradual increase in the number of annotated demonstrations can yield marginal improvements.
Note that \textsc{Skill-KNN} maintains the advantage compared with the raw-input-based selection even with only four annotated demonstrations.
This indicates that the LLM can effectively learn how to rewrite from a limited number of examples and generalize to a wider range of unseen skills.
This generalizability is also supported by our case study in Appendix~\ref{sec:ap_case_study}.

\paragraph{\textsc{Skill-KNN} retains its superiority when the selection of annotation examples is constrained.}
We constrain the selection of annotation examples from two perspectives (detailed in Appendix~\ref{sec:ap_annotation_selection}): first, we limit the SQL operation coverage in annotation examples;
second, we restrict the annotation examples to a few databases.
Figure~\ref{fig:analysis_selection} shows that applying these constraints to the selection of annotation examples leads to only a minor decline in performance, while still maintaining a substantial advantage over raw-input-based selection.

\section{Related Work}

\vspace{2mm}
\noindent \textbf{In-context learning}
has recently become a standard paradigm for effectively leveraging large language models~\citep{brown2020language, hendrycks2020measuring, patel2021mapping, rae2021scaling, zhang2022opt, hoffmann2022training, srivastava2022beyond, chowdhery2022palm, smith2022using, wei2022emergent}.
Such a convenient paradigm has been widely applied in various scenarios such as code generation~\citep{chen2021evaluating, bareiss2022code, li2022competition, chen2023codet, li2023towards}, arithmetic reasoning~\citep{wei2022chain, wang2022self, li2022advance, shi2023language, qin2023chatgpt}, and semantic parsing.
From the view of leveraging skills for in-context learning, most existing work considered explicitly injecting symbolic systems into the response of the model~\citep{cheng2023binding, creswell2023selectioninference, schick2023toolformer, shen2023hugginggpt, lu2023chameleon}.
This work aims to uncover the intrinsic skills from the raw inputs of examples.

\vspace{2mm}
\noindent \textbf{Semantic parsing} with deep learning methods has been explored in much existing work~\citep{dong2016language, yu2018typesql, xu2017sqlnet, guo2019towards, zhong2020grounded, wang2020rat, lin2020bridging, scholak2021picard, qi2022rasat, li2023resdsql}.
Under the recent in-context learning paradigm, there have been some preliminary observations:
\citet{shin2021constrained} showed that GPT-3 is better at generating English-like descriptions rather than the raw logical forms;
\citet{rajkumar2022evaluating} revealed that prompt design is essential for semantic parsing with Codex;
\citet{liu2023comprehensive} showed that ChatGPT has a surprising zero-shot performance on Spider and its variants;
\citet{pourreza2023dinsql} demonstrated that explicitly taking multiple stages for generating SQL leads to better in-context learning performance.
These observations indicate that in-context learning has a great potential on solving semantic parsing tasks,
and this work aims to further activate this potential from the view of improving few-shot selection.

\vspace{2mm}
\noindent \textbf{Few-shot selection} is one essential part for in-context learning.
The standard approach is to use an off-the-shelf embedding model to encode raw inputs and select top-$k$ similar examples~\citep{gao2021making, liu2022makes, hu2022context}.
To improve in semantic parsing tasks, much prior work tried fine-tuning-based methods:
\citet{rubin2022learning} fine-tuned the embedding model based on the conditional probability under the language model;
\citet{poesia2022synchromesh} trained the model to fit the target-side similarity;
\citet{hu2022context} fine-tuned the model based on the similarity of between state changes;
\citet{ye2023compositional} considered the interaction among in-context examples during training the selector.
Our work points in a new direction that does not require further fine-tuning: leveraging task-specific skills by prompting large language models to rewrite input queries.
Beyond semantic parsing, some more recent work tried to explore training-free selection methods for in-context learning from different perspectives:
\citet{nguyen2023context} and \citet{li2023finding} tried to distill the whole example bank into a small set of exemplars and focused on classification tasks;
\citet{wu2023selfadaptive} improved classification tasks through information compression;
\citet{ye2022complementary} and \citet{ye2023explanation} mainly focused explanation-based tasks.
This work proposes prompting extremely large models to facilitate few-shot selection,
which is a novel perspective to harness the power of large language models.

\section{Conclusion}

This work proposes \textsc{Skill-KNN} to facilitate in-context learning on semantic parsing tasks.
By generating skill-based descriptions without any fine-tuning, \textsc{Skill-KNN} and its two variant outperform raw-input-based selections in various tasks.


\section*{Limitations}

\paragraph{GPU resources.}
Our experiments have a high cost on GPU resources, since in-context learning requires extremely large language models.
Specifically, all experiments are conducted on the 8 x NVIDIA A100 GPU station.
During inference time, it takes about 2 x 8 GPU hours to generate for each 10,000 examples.
Thus, it totally takes $400\sim500$ x 8 GPU hours to reproduce our Table~\ref{tab:main_results}.

\paragraph{Task type.}
We mainly evaluate \textsc{Skill-KNN} on cross-domain semantic parsing tasks, and we believe it can also help other challenging tasks where some intrinsic task-specific skills are needed.
However, for tasks that require only surface feature similarity of in-context examples, we suppose the advantage of \textsc{Skill-KNN} could be diminished.

\paragraph{Individual variants.}
We design two variants of \textsc{Skill-KNN} based on consistency and distinctiveness, respectively.
An ideal variant should take into account both these two aspects.
We take this as a future direction for our work.

\section*{Ethics Statement}
Due to the use of pre-trained language models, this work can be exposed to potential ethical risks associated with general deep learning models, such as social bias and privacy breaches.
We suppose this work would be helpful to alleviate potential ethical issues for in-context learning as it can better overcome the surface-form biases in example bank.

\section*{Acknowledgments}
We thank all the anonymous reviewers for their valuable comments.
Shengnan An and Nanning Zheng were supported in part by NSFC under grant No. 62088102.

\bibliography{anthology,custom}
\bibliographystyle{acl_natbib}

\appendix

\clearpage

This is the Appendix of the paper: \textit{
Skill-Based Few-Shot Selection for In-Context Learning}.

\section{More Experimental Results}

We report more experimental results with \textsc{Skill-KNN}.





\subsection{Recall@$N$ Performance of \textsc{Skill-KNN}}

\begin{table}[ht]
\caption{Recall@$N$ performance of \textsc{Skill-KNN} with distinctiveness on Spider dev set.
The backbone model is text-chat-davinci-002 and the sampling temperature is 0.7.
}
\label{tab:ap_topn}
\centering
\resizebox{.99\linewidth}{!}{
\begin{tabular}{@{}c|cccccc@{}}
\toprule
Top-$N$ & 1 & 2 & 3 & 5 & 7 & 10 \\ \midrule
Recall@$N$ (\%) & 80.3 & 85.0 & 87.4 & 89.2 & 89.9 & 90.8 \\ \bottomrule
\end{tabular}
}
\end{table}

Besides the performance of greedy-decoding, we also evaluate the top-$k$ recall performance of \textsc{Skill-KNN}.
Since we cannot take beam search with OpenAI interfaces, we implement the top-$N$ selection with sampling and re-ranking, following the self-consistency setting~\citep{wang2022self}.
Specifically, we first sample 100 sequences and then select the $k$ most frequently occurring sequences and evaluate their execution accuracy.

We take the text-chat-davinci-002 as the backbone model and set the sampling temperature as 0.7.
We evaluate \textsc{Skill-KNN} with distinctiveness on Spider dev set.
Table~\ref{tab:ap_topn} shows that the recall rate gradually goes up with increasing $N$ and can even achieve higher than 90\% when we set $N=10$.
Moreover, the top-$1$ in Table~\ref{tab:ap_topn} is 80.3\% which is higher than the greedy-search performance (78.3\% in Table~\ref{tab:main_results}).
It means that our \textsc{Skill-KNN} can obtain further gains from ensemble methods such as self-consistency.

\subsection{Comparison with More Baseline Methods}~\label{sec:ap_more_compare}

\begin{table}[ht]
\caption{Comparison with more baselines on Spider dev set.
}
\label{tab:ap_more_compare}
\centering
\resizebox{.99\linewidth}{!}{
\begin{tabular}{@{}clc@{}}
\toprule
Backbone & Method & Exec. Acc. \\ \midrule
\multirow{6}{*}{text-chat-davinci-002} & Random & 72.9 \\ 
 & Best-of-5~\citep{nakano2022webgpt} & 73.7 \\
 & K-Center~\citep{sener2018active} & 73.4 \\
 & Influence~\citep{nguyen2023context} & 73.4 \\
 & DDP~\citep{ye2023compositional} & 74.6 \\ \cmidrule(l){2-3} 
 & \textsc{Skill-KNN} w/ SBERT (base) & 76.8 \\ \bottomrule
\end{tabular}
}
\end{table}

Despite the baselines mentioned in Section~\ref{sec:selection_methods}, here we reproduce more selection methods.
Results in Table~\ref{tab:ap_more_compare} further demonstrate the advantage of \textsc{Skill-KNN}.

\section{More Analysis}

\subsection{Motivation Behind Two Variants of \textsc{Skill-KNN}}\label{sec:ap_motivation}

The motivation behind the two variants stems from our consideration of disturbances from the prompt-order sensitivity as additive noises to the ground-truth skills during the rewriting process.
The two variants are designed to address two types of noise:

\paragraph{Zero-mean white noise} which frequently occurs in results and originates from a zero-mean distribution (e.g., zero-mean Gaussian distribution).
We assume its magnitude is relatively small compared to the ground truth.
Zero-mean white noise can cause the loss or redundancy of partial information in the ground truth.

\paragraph{Spike noise} which occasionally occurs in results and has a much larger magnitude than the ground truth.
It strongly influences the information in the ground truth and causes outliers.

\textbf{The consistency-based variant is more effective at addressing zero-mean white noise}, as the averaging operation reduces the variance of zero-mean noise.
\textbf{The distinctiveness-based variant is better suited for handling spike noise}, as it mitigates the influence of outliers.
The final results in our Table 2 indicate that both types of noise occur in our LLM-based rewriting, as evidenced by the close win times of the two variants (19 vs 15).

To further support that two variants are better at tackling two different types of noise, we conduct additional analysis from two perspectives.

\begin{table}[ht]
\caption{The selection accuracy of different variants under different noise patterns. 
}
\label{tab:ap_noise}
\centering
\resizebox{.99\linewidth}{!}{
\begin{tabular}{@{}c|cc@{}}
\toprule
 & Zero-Mean White Noise & Spike Noise \\ \midrule
Consistency & 95.4\% & 81.9\% \\ 
Distinctiveness & 90.4\% & 87.9\% \\ \bottomrule
\end{tabular}
}
\end{table}

\paragraph{Evaluation of selection performance.}
We examined the performance of each variant in selecting better examples from the example bank under the two noise patterns.
We construct some synthetic data for automated evaluation:
we take 1,000 unique embeddings of skill-based descriptions as ground truth, then we add noise on each ground-truth embedding to construct the sample set, and finally we assess whether each sample set could correctly select the original ground-truth embedding.
The accuracies are shown in Table~\ref{tab:ap_noise}.

\paragraph{Human evaluation of rewriting.}
In Spider dev set, we first identified examples where one variant can always succeed with different LLMs while the other variant always failed.
Then, we check the frequency of spike noise occurrences in these examples through human evaluation.
We checked 23 examples and found that:
for examples where the consistency-based variant wins, the spike noise occurs in 0.4 sample/set;
for examples that distinctiveness-based variant wins, the spike noise occurs in 1.6 sample/set.

The results of the above analysis further evidence that the consistency-based variant performs better than the distinctiveness-based variant under the zero-mean white noise but performs worse under the spike noise.

\subsection{The Choice of Hyper-Parameter $m$}\label{sec:ap_m}

\begin{table}[ht]
\caption{Performance with $m=3/5/7/9$ (Dataset: Spider, LLM: code-cushman-002, variant: consistency).
}
\label{tab:ap_m}
\centering
\resizebox{.75\linewidth}{!}{
\begin{tabular}{@{}c|cccc@{}}
\toprule
$m$ & 3 & 5 & 7 & 9 \\ \midrule
Exec. Acc. (\%) & 72.8 & 74.7 & 75.1 & 75.0 \\ \bottomrule
\end{tabular}
}
\end{table}

The two variants of \textsc{Skill-KNN} require to generate $m$ candidate skill-based descriptions.
In our experiments, we set $m=5$ as a trade-off between achieving optimal performance and minimizing computational costs.
During our initial exploration, we had experimented with $m=3/5/7/9$.
As shown in Table~\ref{tab:ap_m}, the performance improves marginally when $m>5$.
Therefore, we decided to set $m=5$.

\subsection{Measuring Diversity of \textsc{Skill-KNN} and Oracle methods}\label{sec:ap_diverse}

\begin{table}[ht]
\caption{Number of different databases among selected examples.
This can reflect the diversity of in-context examples selected by different methods.
}
\label{tab:ap_diversity}
\centering
\resizebox{.99\linewidth}{!}{
\begin{tabular}{@{}c|ccc@{}}
\toprule
Number  of Different Databases & \multicolumn{1}{l}{Spider} & Dr. Spider & \multicolumn{1}{l}{KDBQA} \\ \midrule
\textsc{Skill-KNN} (distinctiveness) & 2.83 & 2.84 & 2.88 \\
Target-KNN (oracle) & 2.18 & 2.16 & 2.21 \\ \bottomrule
\end{tabular}
}
\end{table}

A surprising observation in Table~\ref{tab:main_results} is that with the same backbone model for in-context learning, our \textsc{Skill-KNN} could sometimes outperforms oracle methods.
Specifically, \textsc{Skill-KNN} consistently outperforms at least one oracle method on the DB sub-task in Dr. Spider and two sub-tasks in COGS (marked with underlines).
Such an observation could be caused by the different diversity in selected examples.
As indicated in \citet{levy2022diverse} and \citet{an2023incontext}, beyond the similarity to the test case, a higher diversity among in-context examples could also help better perform cross-domain generalization under in-context learning.
Oracle methods directly seek higher similarity, thus the selected examples may be less diverse, which could slightly hamper the generalization performance.
To reflect the diversity of in-context examples, we count the number of different databases among selected examples.
Statistics in Table~\ref{tab:ap_diversity} shows that \textsc{Skill-KNN} can lead to a higher diversity than oracle method, which is in line with our hypothesis.

\begin{figure*}[ht]
	\centering
	\subfloat[Raw-Input-Based Embedding Space\label{fig:surface_form_embed}
]{\includegraphics[width=.45\linewidth]{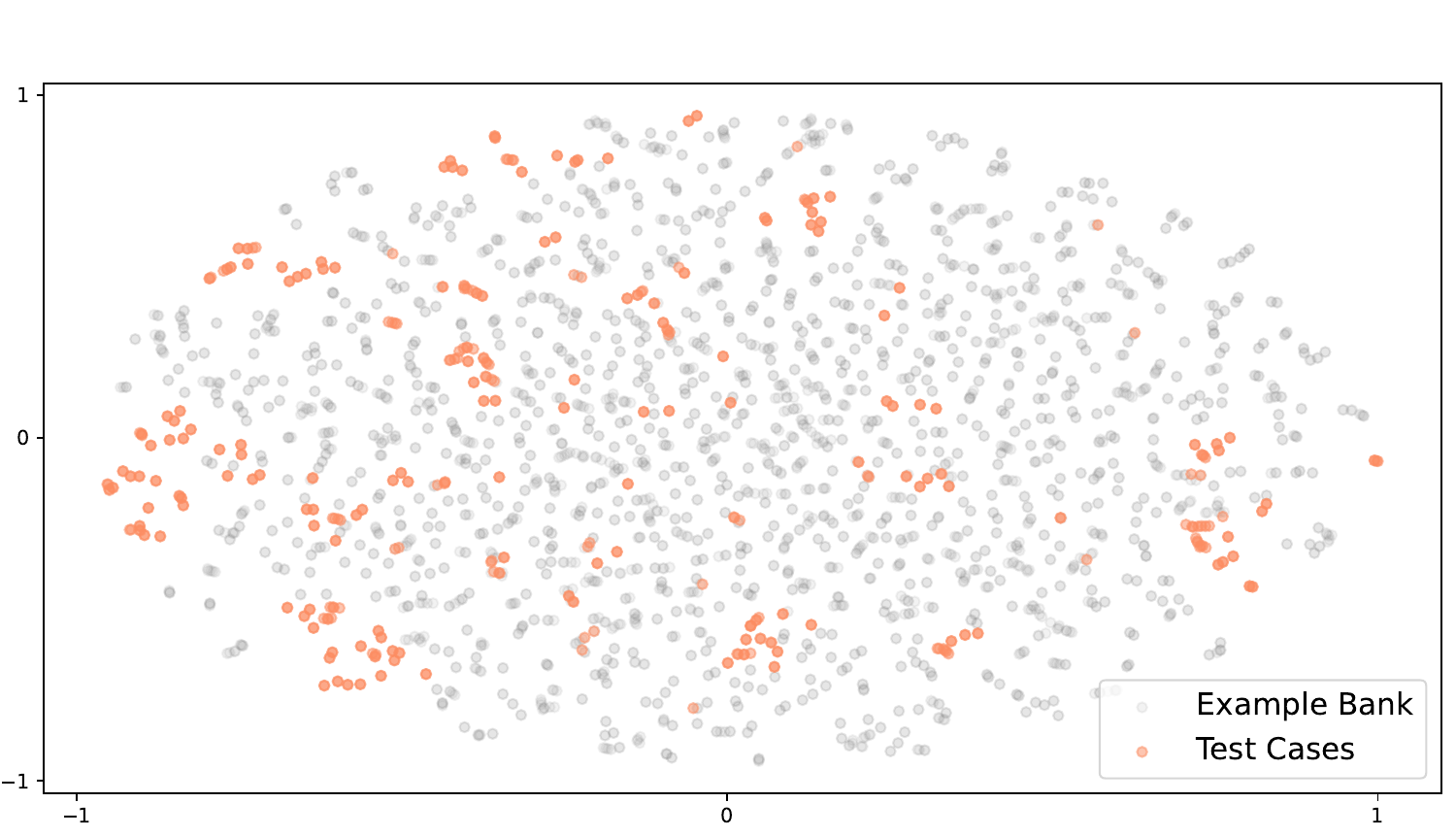}}\hfill
	\subfloat[Skill-Based Embedding Space\label{fig:skill_embed}]{\includegraphics[width=.45\linewidth]{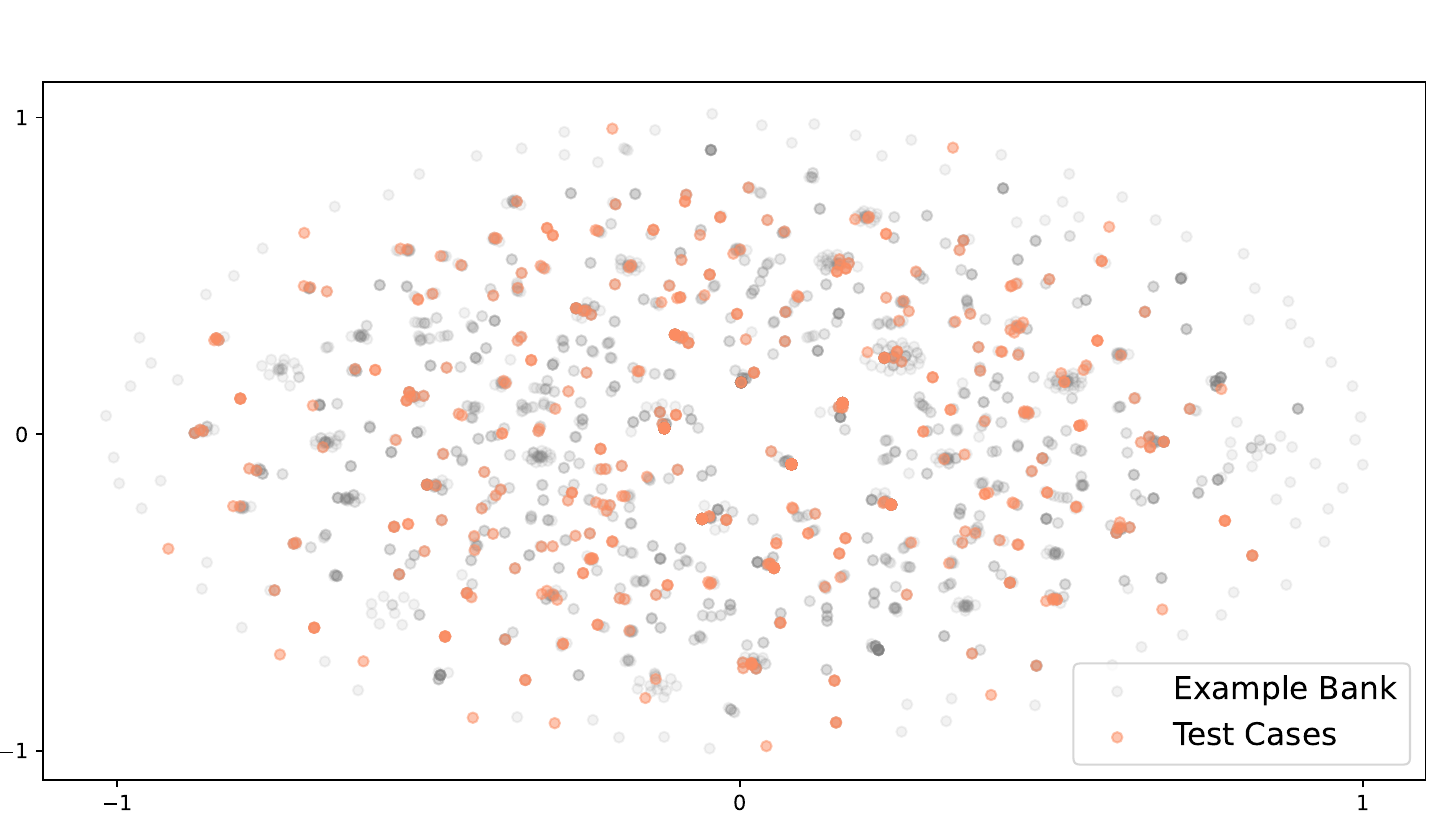}}
	\caption{
    T-SNE visualization for the embedding space of (a) raw input queries and (b) skill-based descriptions.
    The orange points are from the dev set in Spider while the gray points are from example bank.
    }\label{fig:visualization}
\end{figure*}

\subsection{Why do skill-based descriptions perform better?}

Since both \textsc{Skill-KNN} and raw-input-based methods use the embedding similarity for selection, we suppose that the higher performance of \textsc{Skill-KNN} can be contributed by some desired properties in the embedding space of skill-based descriptions.
Based on this inspiration, we visualize the embedding space of both raw input queries and skill-based descriptions with t-SNE~\citep{van2008visualizing}.
More details are contained in Appendix~\ref{sec:ap_tsne}.

\paragraph{Under the embedding space of skill-based descriptions, the distribution of test cases is closer to that of the example bank, thus benefiting cross-domain generalization.}
For the raw-input-based embeddings of test cases, Figure~\ref{fig:surface_form_embed} shows that these embeddings are mainly centralized in some local parts.
On the one hand, the example bank can not be fully utilized under this embedding space, since the top-$k$ similar examples must be around the local parts of test cases.
On the other hand, the different distributions represent that this space does not reveal the inner similarity between test cases and example bank, thus is helpless to facilitate cross-domain generalization.
Under the skill-based embedding space (shown in Figure~\ref{fig:skill_embed}), the distributions of test cases and the example bank are better matched.
Therefore, the cross-domain generalization gap can be better bridged with skill-based descriptions.

\paragraph{The skill-based embedding space is more sparse, thus the boundary of similar examples is more clear and can be more robust to perturbations.}
As shown in Figure~\ref{fig:surface_form_embed}, the raw-input-based embedding space is almost evenly distributed.
It means that for one embedding in this space, the boundary to determine "which examples are similar" is not clear enough.
Without a clear boundary for selecting similar examples, the performance could be non-robust to perturbations in data.
Compared with the embeddings of raw inputs, the skill-based embeddings shown in Figure~\ref{fig:skill_embed} are more clustered, thus the KNN-based selection can be more robust to data perturbations.

\begin{figure*}[ht]
	\centering
	\subfloat[Number of Tables\label{fig:table_numbers}
]{\includegraphics[width=.45\linewidth]{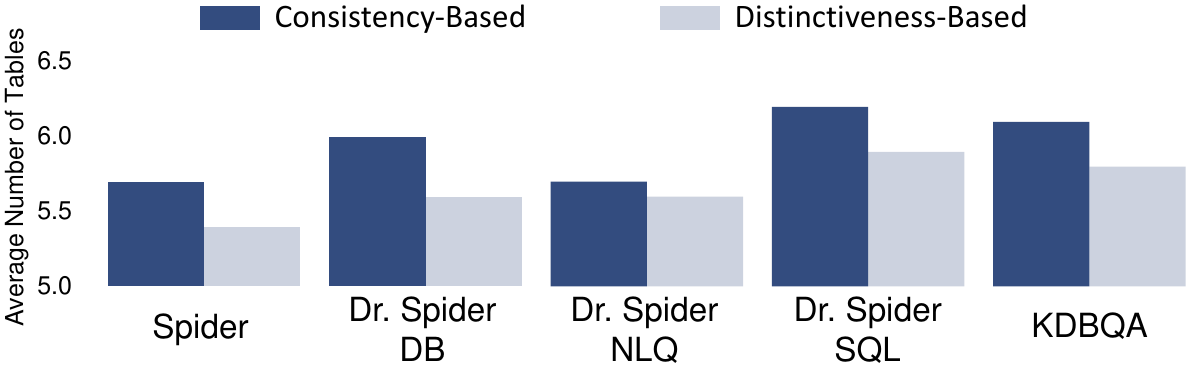}}\hfill
	\subfloat[Length of SQL Queries\label{fig:sql_length}]{\includegraphics[width=.45\linewidth]{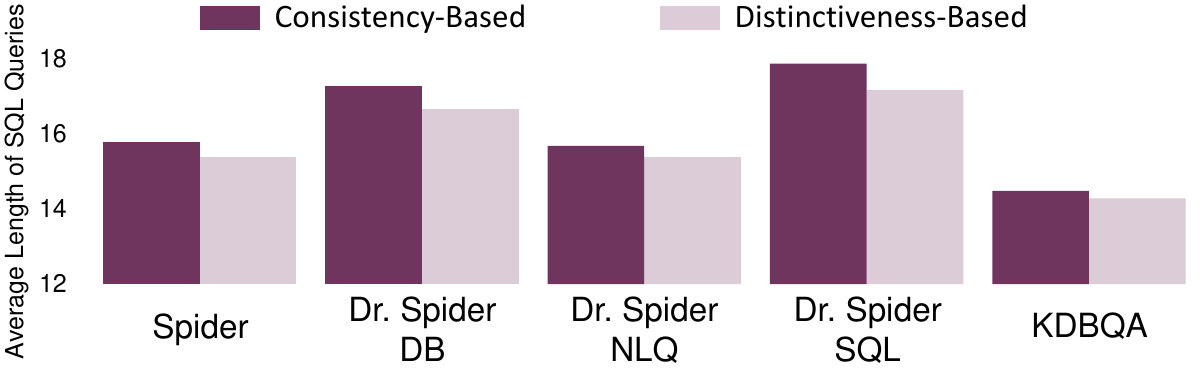}}
	\caption{
    Two perspectives that reflect the complexity of selected in-context examples from two variants of \textsc{Skill-KNN}.
    (a) The average number of tables contained in the database of each in-context example.
    (b) The average length of SQL queries (split by space).
    }\label{fig:complexity}
\end{figure*}

\subsection{What factors cause the different performances of two variants?}\label{sec:analysis_two_enhance}

As mentioned in Section~\ref{sec:main_results}, different backbone models prefer different variants of \textsc{Skill-KNN}.
As both two methods aim to improve skill-based similarity, such performance differences indicate that beyond similarity, some other factors also influence in-context learning.
\citet{an2023incontext} and \citet{rajkumar2022evaluating} indicated that the complexity of selected examples could be one potential factor.
Here, we explore whether the two variants differ in the complexity of the selected examples.

\paragraph{Consistency-based variant always leads to more complex in-context examples than distinctiveness-based variant.}
Here, we mainly check the complexity of in-context examples for text-to-SQL tasks from two perspectives:
(1) the average number of tables, which reflects the complexity of database, and (2) the average length of SQL queries, which reflects the hardness of searching.
As shown in Figure~\ref{fig:complexity}, consistency-based variant leads to higher complexity than distinctiveness-based variant in all tasks and under both perspectives.
It can help to explain why different models prefer different variants:
text-chat-davinci-002 and code-davinci-002 prefer simpler in-context examples
while text-davinci-003 and code-cushman-002 are more robust to different complexity.

\section{Case Study}\label{sec:ap_case_study}

\begin{table*}[t]
\caption{Case study for generated skill-based descriptions in Spider.
Texts in \textcolor[HTML]{8409E9}{purple} are novel descriptions that are not presented in our annotations.
}
\label{tab:ap_case_study}
\centering
\resizebox{.99\linewidth}{!}{
\begin{tabular}{@{}lll@{}}
\toprule
Input Query & Generated Skill & SQL \\ \midrule
\begin{tabular}[c]{@{}l@{}}display the emails of the employees who have no commission percentage and salary \\ within the range 7000 to 12000 and works in that department which number is 50.\end{tabular} & \begin{tabular}[c]{@{}l@{}}To solve this task in the database, we need to select one column, apply an \\ interval constraint, \textcolor[HTML]{8409E9}{apply a null constraint}, and apply an equality constraint.\end{tabular} & \begin{tabular}[c]{@{}l@{}}SELECT email FROM employees WHERE commission\_pct  =  "null" AND \\ salary BETWEEN 7000 AND 12000 AND department\_id = 50\end{tabular} \\ \midrule
What are the salaries and manager ids for employees who have managers? & \begin{tabular}[c]{@{}l@{}}To solve this task in the database, we need to select two columns and apply \\ \textcolor[HTML]{8409E9}{a non-null constraint on the values in another column}.\end{tabular} & SELECT salary ,  manager\_id FROM employees WHERE manager\_id != "null" \\ \midrule
What is all the information about the Marketing department? & \begin{tabular}[c]{@{}l@{}}To solve this task in the database, we need to \textcolor[HTML]{8409E9}{select all columns} and apply \\ a constraint on the values in one column.\end{tabular} & SELECT * FROM  departments WHERE department\_name  =  'Marketing' \\ \bottomrule
\end{tabular}
}
\end{table*}

Considering that the quality of generated skill-based description can be one key factor that influences the effectiveness of \textsc{Skill-KNN}, we manually check the generated skills for 100 examples.
We find that 86/100 generated skills are exactly correct;
12/100 are almost correct but need some partial modifications (e.g., the number of joined tables);
and only 2/100 generated skills are totally wrong.

Moreover, during manually checking the quality of generated skills, we surprisingly find that there are some novel descriptions about skills that are not presented in our annotated examples.
Table~\ref{tab:ap_case_study} shows some examples.
It indicates that the prompting-based rewriting can provide a degree of generalization of unannotated skills.

\section{Detailed Settings of Experiments}\label{sec:ap_settings}

In this section, we provide more details about our experimental settings.

\subsection{Select Examples for Annotation}\label{sec:ap_annotation_selection}

In our default setting, we consider two principles to select examples for annotating the required skills:
1) ensuring coverage of all logical operations found in the example bank and 2) selecting examples from diverse databases.
Specifically, we first find all used logical operations in the example bank and greedily cover these operations in a few examples.
Then, we randomly select more examples from various databases until there are 16 examples.

In our ablation study shown in Figure~\ref{fig:analysis_selection}, to constrain the operation coverage, we just remove our first selection step;
to constrain the database diversity, we just select examples from two databases.

\subsection{Inference Hyper-Parameters}

During inference, we set the max decoding length to 200, and the sampling temperature to 0.

\subsection{Input-Output Formats}\label{sec:ap_input_output_formats}

\begin{figure*}[ht]
	\centering
	\subfloat[Formats of Text-to-SQL Tasks (with Optional Evidence for BIRD)\label{fig:ap_format_text2sql}
]{\includegraphics[width=.9\linewidth]{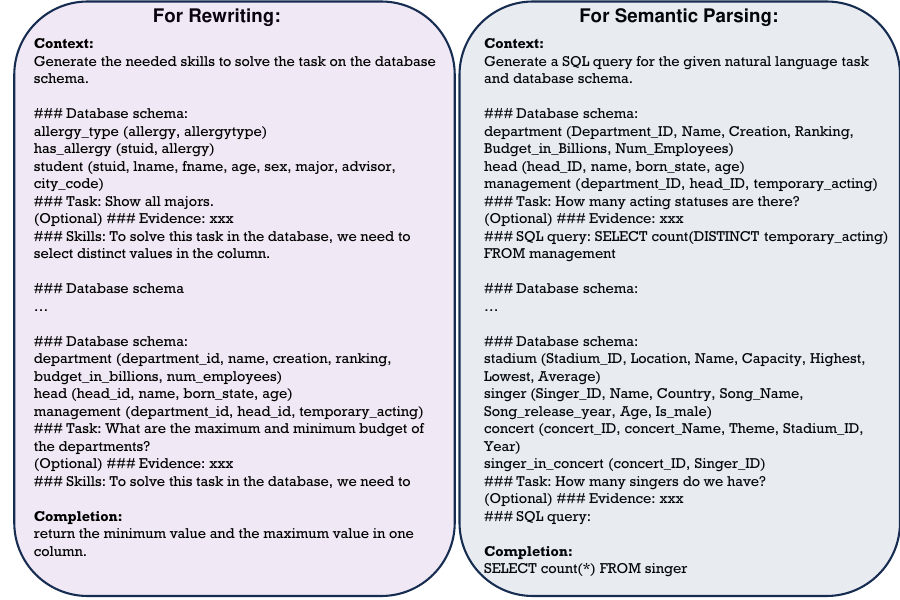}}\\
	\subfloat[Formats of COGS\label{fig:ap_format_cogs}]{\includegraphics[width=.9\linewidth]{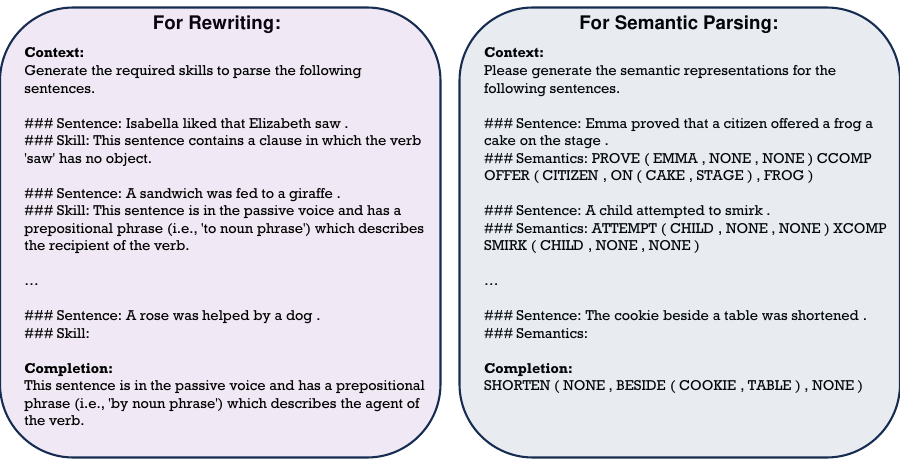}}
	\caption{Input-output formats used in our experiments.
    }\label{ap:input_output_formats}
\end{figure*}

Figure~\ref{ap:input_output_formats} shows some input-output examples to illustrate the data formats in our experiments.

Note that the output format of COGS follows the transformation in \citet{an2023incontext} which converts the original long-chain format into a more compact function-calling format.
Such a transformation is similar to the conversion from Lambda calculus to FunQL in Geo domain\citep{zelle1996learning, kate2005learning, zettlemoyer2012learning}.
It improves the human readability by omitting two types of details in original format: the special marker for definite descriptions and the Skolem constants.
Apart from the omitted details, this transformation keeps the main semantics in the domain of COGS, such as semantic roles, modifications, and orders among clauses and modifications.

\subsection{Evaluation on BIRD}\label{sec:ap_bird}
Different from other text-to-SQL tasks, BIRD additionally provides ``evidence'' for each natural language question.
Therefore, we add the evidence as part of the context for in-context learning.
For evaluating raw-input-based methods on BIRD, we concatenate the natural language question and the additional evidence to compute the embedding.
For our \textsc{Skill-KNN}, we also provide the evidence for rewriting, and we use 12 annotated demonstrations with evidence (shown in Appendix~\ref{sec:ap_annotated_examples}).

Since the database schema in BIRD is too large to be fully contained in the context for LLM, we reduce the size of schema through grounding in pre-processing.
Specifically, we calculate the embedding similarity between the input question (along with the evidence) and each table name and column name.
Based on this similarity, we preserve 8 tables each with 16 columns for each schema-question pair.

\subsection{Evaluation on COGS}
COGS totally contains 24,155 examples in train set and 21,000 examples in gen set.
To reduce the high computational cost, we sample 2,000 examples from the train set as the example bank for in-context learning, and sample 1,000 examples from two sub-tasks primitive substitution (P.S.) and primitive structural alternation (P.A.) which are defined in \citet{an2023incontext}.

\subsection{Target Sketch Matching for SQL}\label{sec:ap_target_sketch_match_sql}
As mentioned in Section~\ref{sec:selection_methods}, to select in-context examples with target sketch matching (oracle) in text-to-SQL tasks, we calculate the overlap of SQL key words between the example from example bank and the labeled SQL query of the test input query.
We mainly consider the following SQL key words along with several operations for calculation:
\texttt{SELECT, WHERE, GROUP, HAVING, ORDER, DESC, ASC, LIMIT, JOIN, INTERSECT, EXCEPT, UNION, NOT, IN, OR, AND, BETWEEN, EXISTS, LIKE, DISTINCT, COUNT, AVG, MIN, MAX, SUM, CAST, CASE, WHEN, THEN, ELSE, END, IIF, REAL, FLOAT, NULL, STRFTIME, *, /, =, >, ,<, !, +, -, \%}.
Based on these key words, the target sketch similarity between two SQL queries $y_t$ and $y_i$ is calculated as follows,
\begin{equation}
    \mathrm{sim}_k (y_t, y_i) = |\mathrm{KW}(y_t)\cap\mathrm{KW}(y_i)|,
\end{equation}
in which $\mathrm{KW}(\cdot)$ returns a set of contained key words.

\subsection{T-SNE Visualization}\label{sec:ap_tsne}
For the visualized embedding space in Figure~\ref{fig:visualization}, we use Sentence-BERT as the embedding model and take examples from both example bank and dev set in Spider.
For \textsc{Skill-KNN}, we take its consistency-based variant.
To accelerate the visualization process, we just take examples with medium hardness (defined in \citet{yu2018spider}).
We use the implementation of t-SNE from the sklearn library\footnote{\url{https://scikit-learn.org/stable/}.}.
We set the learning rate of t-SNE as ``auto'', init method as ``random'', and perplexity as 3.

\section{Annotated Demonstrations}\label{sec:ap_annotated_examples}

Table~\ref{tab:ap_annotated_text2sql} lists 16 annotated demonstrations for text-to-SQL tasks and
Table~\ref{tab:ap_annotated_bird} lists another 12 annotated demonstrations with evidence (which is required in BIRD).
Table~\ref{tab:ap_annotated_cogs} lists 16 annotated demonstrations for COGS.
Appendix~\ref{sec:ap_annotation_selection} introduces how we select these examples.

\begin{table*}[ht]
\caption{Annotated demonstrations for text-to-SQL tasks.
All these examples are from the example bank of Spider.
}
\label{tab:ap_annotated_text2sql}
\centering
\resizebox{.99\linewidth}{!}{
\begin{tabular}{@{}lll@{}}
\toprule
DB ID & Input Query & Annotated Skill-Based Descriptions \\ \midrule
allergy\_1 & Show all majors. & To solve this task in the database, we need to select distinct values in   the column. \\ \midrule
match\_season & \begin{tabular}[c]{@{}l@{}}Count the number of different colleges that players \\ who play for Columbus   Crew are from.\end{tabular} & \begin{tabular}[c]{@{}l@{}}To solve this task in the database, we need to join two tables and count   the number \\ of distinct values in the column.\end{tabular} \\ \midrule
gymnast & \begin{tabular}[c]{@{}l@{}}What are the hometowns of gymnasts and the \\ corresponding number of   gymnasts?\end{tabular} & \begin{tabular}[c]{@{}l@{}}To solve this task in the database, we need to join two tables, select   one column, \\ group these selections and count the number of selections in each   group.\end{tabular} \\ \midrule
gas\_company & \begin{tabular}[c]{@{}l@{}}Show minimum, maximum, and average market \\ value for all companies.\end{tabular} & \begin{tabular}[c]{@{}l@{}}To solve this task in the database, we need to return the minimum value,   the \\ maximum value, and the average of values in the column.\end{tabular} \\ \midrule
culture\_company & \begin{tabular}[c]{@{}l@{}}Show the years, book titles, and publishers for all books, \\ in descending   order by year.\end{tabular} & \begin{tabular}[c]{@{}l@{}}To solve this task in the database, we need to select three columns and   sort them \\ in descending order according to the values in one column.\end{tabular} \\ \midrule
product\_catalog & \begin{tabular}[c]{@{}l@{}}Which catalog contents have a product stock number \\ that starts from   "2"? Show the catalog entry names.\end{tabular} & \begin{tabular}[c]{@{}l@{}}To solve this task in the database, we need to select one column and   apply a \\ constraint on the format of values in this column.\end{tabular} \\ \midrule
bike\_1 & \begin{tabular}[c]{@{}l@{}}What are the dates in which the mean sea level pressure \\ was between 30.3   and 31?\end{tabular} & \begin{tabular}[c]{@{}l@{}}To solve this task in the database, we need to select one column and   apply a \\ constraint that the values in another column should in a certain   range.\end{tabular} \\ \midrule
flight\_1 & \begin{tabular}[c]{@{}l@{}}What is the salary and name of the employee who has \\ the most number of   certificates on aircrafts with distance \\ more than 5000?\end{tabular} & \begin{tabular}[c]{@{}l@{}}To solve this task on the database, we need to join three tables, apply a   \\ greater-than constraint, group the selections and calculate the number of   each group, \\ sort the selections in descending order, and select the top   result.\end{tabular} \\ \midrule
bike\_1 & \begin{tabular}[c]{@{}l@{}}What is the average longitude of stations that never had \\ bike   availability more than 10?\end{tabular} & \begin{tabular}[c]{@{}l@{}}To solve this task in the database, we need to calculate the average   value in one \\ colum, apply an non-inclusion constraint with another set of   selections, which \\ need to group the selctions and find which groups have a   maximum value greater \\ than the threshold.\end{tabular} \\ \midrule
hr\_1 & \begin{tabular}[c]{@{}l@{}}display all the information of employees whose salary is \\ in the range of   8000 and 12000 and commission is not null \\ or department number does not equal   to 40.\end{tabular} & \begin{tabular}[c]{@{}l@{}}To solve this task in the database, we need to give full information   about selections, \\ apply an interval constraint, and apply an optional   constraint that two unequal \\ judgments should be satisfied at least one.\end{tabular} \\ \midrule
bike\_1 & \begin{tabular}[c]{@{}l@{}}What are the ids of stations that have latitude above 37.4 \\ and never had   bike availability below 7?\end{tabular} & \begin{tabular}[c]{@{}l@{}}To solve this task in the database, we need to exclude the selections in   the second set \\ from the first set: the first set of selections need to apply   a greater-than constraint, and \\ the second set of selctions need to group the   selctions and find which groups have a \\ minimum value lower than the   threshold.\end{tabular} \\ \midrule
bike\_1 & \begin{tabular}[c]{@{}l@{}}What are the names and ids of stations that had more \\ than 14 bikes   available on average or were installed \\ in December?\end{tabular} & \begin{tabular}[c]{@{}l@{}}To solve this task in the database, we need to return the union of two   set of selections: \\ the first set of selections need to join two tables, group   the selections and find which \\ groups have an average value greater than the   threshold, and the second set of selections \\ need to apply a constaint on the   format of values.\end{tabular} \\ \midrule
storm\_record & \begin{tabular}[c]{@{}l@{}}Show storm name with at least two regions and \\ 10 cities affected.\end{tabular} & \begin{tabular}[c]{@{}l@{}}To solve this task in the database, we need to return the intersection of   two set of \\ selections: the first set of selections need to join two tables,   group the selections and \\ find which groups have a number of selections   greater than or equal to the threshold, \\ and the second set of selections need   to join two tables, group the selections \\ and find which groups have a sum of   values larger than or equal to the threshold.\end{tabular} \\ \midrule
formula\_1 & \begin{tabular}[c]{@{}l@{}}List the forenames of all distinct drivers in \\ alphabetical order?\end{tabular} & \begin{tabular}[c]{@{}l@{}}To solve this task in the database, we need to select distinct values in   one column \\ and sort these selections in ascending order according to the   selected values.\end{tabular} \\ \midrule
hr\_1 & \begin{tabular}[c]{@{}l@{}}display job ID for those jobs that were done by two \\ or more for more than   300 days.\end{tabular} & \begin{tabular}[c]{@{}l@{}}To solve this task in the database, we need to apply a greater-than   constraint on \\ the difference between two values, group the selections and   find which groups have \\ a number of selections greater than or equal to the   threshold.\end{tabular} \\ \midrule
small\_bank\_1 & \begin{tabular}[c]{@{}l@{}}Find the names and total checking and savings balances \\ of accounts whose   savings balance is higher than the \\ average savings balance.\end{tabular} & \begin{tabular}[c]{@{}l@{}}To solve this task in the database, we need to select one column and add   the values \\ in another two columns, join three tables, and apply a   greater-than constraint where \\ the threshold is the average of another set of   selected values.\end{tabular} \\ \bottomrule
\end{tabular}
}
\end{table*}

\begin{table*}[ht]
\caption{Annotated demonstrations for text-to-SQL task with evidence.
All these examples are from the example bank of BIRD.
}
\label{tab:ap_annotated_bird}
\centering
\resizebox{.99\linewidth}{!}{
\begin{tabular}{@{}llll@{}}
\toprule
DB ID & Input Query & Evidence & Annotated Skill-Oriented Descriptions \\ \midrule
superstore & \begin{tabular}[c]{@{}l@{}}Please list any three orders that caused a loss \\ to the company.\end{tabular} & caused a loss to the company refers to Profit < 0 & \begin{tabular}[c]{@{}l@{}}To solve this task in the database, we need to select one column, \\ apply a   less-than constraint, and return three results.\end{tabular} \\ \midrule
disney & \begin{tabular}[c]{@{}l@{}}Calculate the percentage of voice actors whose \\ main character in the   movie is in the Drama genre.\end{tabular} & \begin{tabular}[c]{@{}l@{}}DIVIDE(COUNT(voice-actor where genre = 'Drama'), \\ COUNT(voice-actor)) as   percentage;\end{tabular} & \begin{tabular}[c]{@{}l@{}}To solve this task in the database, we need to join three tables and   \\ calculate a percentage number. Additionally, to calculate the percentage  \\  number, we need to count the number of values with an equivalent \\ constraint,   cast the count into a real number, multiply this number \\ by 100, and divide it   by the another count.\end{tabular} \\ \midrule
legislator & \begin{tabular}[c]{@{}l@{}}Among the legislators who will end in 2009, \\ how many are from the   Republican party?\end{tabular} & \begin{tabular}[c]{@{}l@{}}the legislators who will end in 2009 refers to END 2009; \\ from the   Republican party refers to party = 'Republican'\end{tabular} & \begin{tabular}[c]{@{}l@{}}To solve this task in the database, we need to select two columns, \\ apply   an equivalent constraint on the time and an equivalent \\ constraint on the text   value.\end{tabular} \\ \midrule
works\_cycles & \begin{tabular}[c]{@{}l@{}}Calculate the average length of employment for \\ employee working in the   Research and Development \\ deparment.\end{tabular} & \begin{tabular}[c]{@{}l@{}}average length of employment = \\ AVG(subtract(2022, year(HireDate)))\end{tabular} & \begin{tabular}[c]{@{}l@{}}To solve this task in the database, we need to get two times, calculate   \\ the differences between two times, and calculate the average of these   \\ differences. Additionally, we need to join three tables and apply an   \\ equivalent constraint.\end{tabular} \\ \midrule
retail\_complains & \begin{tabular}[c]{@{}l@{}}Among the teenager clients who use Google \\ account and Microsoft account,   which group \\ of client is more than the other?\end{tabular} & \begin{tabular}[c]{@{}l@{}}teenager refers to 13 < age < = 19; Google account refers \\ to email   like '\%@gmail.com'; Microsoft account refers \\ to email like '\%@outlook.com'\end{tabular} & \begin{tabular}[c]{@{}l@{}}To solve this task in the database, we need to compare the number of   \\ values in two formats, and return the value that has a higher number.   \\ Additionally, we need to apply a between-and constraint.\end{tabular} \\ \midrule
university & \begin{tabular}[c]{@{}l@{}}What are the top three universities with the most \\ international students?\end{tabular} & \begin{tabular}[c]{@{}l@{}}most international students refers to   MAX(SUM(DIVIDE(\\ MULTIPLE(pct\_international\_students, num\_students), 100)));  \\  name of university refers to university\_name;\end{tabular} & \begin{tabular}[c]{@{}l@{}}To solve this task in the database, we need to select distinct values   \\ from one cloumn, join two tables, group the results, order the results \\ in   descending order according to the sum of values, and return the top three   \\ results. Additionally, during getting the sum of values, we need to multiply   \\ the values in one column with percentages in another column and divide the  \\  results by 100.\end{tabular} \\ \midrule
airline & \begin{tabular}[c]{@{}l@{}}What is the percentage of flights which landed \\ at Pittsburgh were faster   than scheduled?\end{tabular} & \begin{tabular}[c]{@{}l@{}}percentage = MULTIPLY(DIVIDE(SUM(\\ ACTUAL\_ELAPSED\_TIME <   T2.CRS\_ELAPSED\_TIME), \\ COUNT(Code)), 100); landed at refers to DEST;   \\ Pittsburgh refers to Description which contains 'Pittsburgh'; \\ faster than   scheduled refers to \\ ACTUAL\_ELAPSED\_TIME < CRS\_ELAPSED\_TIME;\end{tabular} & \begin{tabular}[c]{@{}l@{}}To solve this task in the database, we need to compare values in two   \\ columns and convert the comparison result into a percentage. Additionally, \\ we   need to join two tables, apply a constraint on the format of values in one   \\ column, and ensure that the values in two columns are not null. Moreover, to  \\ get the percentage number, we need to cast the sum of values into a real   \\ number, multiply this number by 100, and divide it by the total count.\end{tabular} \\ \midrule
retail\_complains & \begin{tabular}[c]{@{}l@{}}Between 1/1/2017 and 4/1/2017, what is the average \\ server time of calls   under the server DARMON?\end{tabular} & \begin{tabular}[c]{@{}l@{}}between 1/1/2017 and 4/1/2017 refers to Date received between  \\  '2017-01-01' and '2017-04-01'; \\ average server time refers to avg(ser\_time)\end{tabular} & \begin{tabular}[c]{@{}l@{}}To solve this task in the database, we need to calculate the average of   \\ the selected values and apply a between-and constraint. Additionally, to   \\ obtain the times from values in text format, we need to extract substrings   \\ from these texts and cast them into real numbers.\end{tabular} \\ \midrule
soccer\_2016 & When did Chennai Super Kings play its first match? & \begin{tabular}[c]{@{}l@{}}match date refers to Match\_Date; Chennai Super Kings refers \\ to Team\_Name   = 'Chennai Super Kings'; \\ first match refers to min(Match\_Date)\end{tabular} & \begin{tabular}[c]{@{}l@{}}To solve this task in the database, we need to apply an either-or   \\ constraint, sort the selected results in ascending order, and return the top   \\ one result.\end{tabular} \\ \midrule
retails & \begin{tabular}[c]{@{}l@{}}Which ship mode has more "deliver in person" \\ instructions, rail   or mail?\end{tabular} & \begin{tabular}[c]{@{}l@{}}ship mode refers to l\_shipmode; "deliver in person" instruction   \\ refers to l\_shipinstruct = 'DELIVER IN PERSON'\end{tabular} & \begin{tabular}[c]{@{}l@{}}To solve this task in the database, we need to count the number of two   \\ values in ome column and return the value with a larger count. Additionally,   \\ we need to apply an equality constraint.\end{tabular} \\ \midrule
cookbook & \begin{tabular}[c]{@{}l@{}}Which ingredient appeared the most in recipes? \\ Calculate its amount of   appearance in percentage.\end{tabular} & \begin{tabular}[c]{@{}l@{}}ingredient appeared the most in recipes refers to   MAX(\\ COUNT(ingredient\_id)); calculation =   MULTIPLY(DIVIDE(\\ COUNT(MAX(ingredient\_id)), COUNT(ingredient\_id)), 100)\end{tabular} & \begin{tabular}[c]{@{}l@{}}To solve this task in the database, we need to select one column and   \\ calculate one percentage number, join two tables, group the selected results,   \\ and sort the results in descending order according to the size of each group.   \\ Additionally, to calculate the percentage number, we need to cast the count   \\ of vaules into a float number, multiply this number by 100, and divide it by   \\ the another count.\end{tabular} \\ \bottomrule
\end{tabular}
}
\end{table*}

\begin{table*}[ht]
\caption{Annotated demonstrations for COGS.
All these examples are from the example bank of COGS.
}
\label{tab:ap_annotated_cogs}
\centering
\resizebox{.99\linewidth}{!}{
\begin{tabular}{@{}ll@{}}
\toprule
Input Query & Annotated Skill-Based Descriptions \\ \midrule
Isabella liked that Elizabeth saw . & This sentence contains a clause in which the verb 'saw' has no object. \\ \midrule
A sandwich was fed to a giraffe . & \begin{tabular}[c]{@{}l@{}}This sentence is in the passive voice and has a prepositional phrase   \\ (i.e., 'to noun phrase') which describes the recipient of the verb.\end{tabular} \\ \midrule
Benjamin froze . & The verb 'froze' has no object. \\ \midrule
Sophia was given a cookie by Emma . & \begin{tabular}[c]{@{}l@{}}This sentence is in the passive voice, has an object and has a   prepositional \\ phrase (i.e., 'by noun phrase') which describes the agent of   the verb.\end{tabular} \\ \midrule
Sophia liked a box on the cake . & This sentence has a single object with a modification phrase. \\ \midrule
Emma sold the drink beside a road to a   zebra . & \begin{tabular}[c]{@{}l@{}}This sentence has a direct object with a modification phrase and has a   prepositional \\ phrase (i.e., 'to noun phrase') which describes the recipient   of th verb.\end{tabular} \\ \midrule
A lion ate . & The verb 'ate' has no object. \\ \midrule
Eleanor was offered the ball . & This sentence is in the passive voice and has an object. \\ \midrule
A box was helped . & This sentence is in the passive voice and has no object or prepositional   phrase. \\ \midrule
The fish dreamed to walk . & This sentence has an infinitive verb. \\ \midrule
A cat lended a lawyer the cake . & This sentence has an indirect object and a direct object. \\ \midrule
The basket was handed to a cat by Emma . & \begin{tabular}[c]{@{}l@{}}This sentence is in the passive voice and has two prepositional phrases:   the first one \\ (i.e., 'to noun phrase') describes the recipient of the verb   and the second one (i.e., 'by noun phrase') \\ describes the agent of the verb.\end{tabular} \\ \midrule
Sofia thought that a pancake rolled . & This sentence contains a clause in which the verb 'rolled' has no object. \\ \midrule
Liam hoped that the boy wanted to dance . & This sentence contains a clause that has an infinitive verb. \\ \midrule
The cat gave Ethan a rose on the table . & This sentence has an indirect object and a direct object with a   modification phrase. \\ \midrule
\begin{tabular}[c]{@{}l@{}}The duke was passed the shell on a table   \\ in the house by Emma .\end{tabular} & \begin{tabular}[c]{@{}l@{}}This sentence is in the passive voice, has an object with a nested   modification phrase, \\ and has a prepositional phrase (i.e., 'by noun phrase')   which describes the agent of the verb.\end{tabular} \\ \bottomrule
\end{tabular}
}
\end{table*}

\end{document}